\documentclass[11pt]{article}

\usepackage[T1]{fontenc}
\usepackage[utf8]{inputenc}
\usepackage[a4paper,margin=1in]{geometry}
\usepackage{lmodern}
\usepackage{microtype}

\usepackage{amsmath,amssymb}
\newtheorem{theorem}{Theorem}[section]

\usepackage{graphicx}
\graphicspath{{./}{figures/}{}}
\DeclareGraphicsExtensions{.pdf,.png,.jpg}

\usepackage{xcolor}
\definecolor{joBlue}{RGB}{27,74,144}

\usepackage{tikz}
\usetikzlibrary{arrows.meta,positioning,calc}

\usepackage{float}
\usepackage[section]{placeins}
\usepackage{placeins}
\FloatBarrier

\usepackage{booktabs}
\usepackage{siunitx}
\usepackage{booktabs,tabularx,makecell,array}
\usepackage{ragged2e}
\newcolumntype{Y}{>{\RaggedRight\arraybackslash}X}

\usepackage{adjustbox}
\usepackage{caption}
\usepackage{subcaption}
\usepackage{enumitem}
\usepackage[most]{tcolorbox}
\usepackage{multirow}
\usepackage{mathtools}
\usepackage[percent]{overpic}
\usepackage{algorithm}
\usepackage{algpseudocode}

\usepackage[hyphens]{url}
\usepackage{xurl}
\usepackage[hidelinks]{hyperref}
\hypersetup{breaklinks=true}
\usepackage[numbers,sort&compress]{natbib}

\title{KD-PINN: Knowledge-Distilled PINNs for ultra-low-latency real-time neural PDE solvers}

\author{
  Karim Bounja$^{1}$\thanks{Corresponding author. Email: k.bounja.doc@uhp.ac.ma}
  \and
  Lahcen Laayouni$^{2}$
  \and
  Abdeljalil Sakat$^{1}$
}

\date{} 

\begin{document}

\maketitle

\begin{center}
\small
$^{1}$Laboratory for Analysis and Modeling of Systems and Decision Support (LAMSAD),\\
National School of Applied Sciences, Hassan 1st University of Settat,\\
Berrechid 26100, Morocco\\[4pt]
$^{2}$Department of Computer Science, School of Science and Engineering,\\
Al Akhawayn University in Ifrane,\\
Ifrane 53000, Morocco
\end{center}

\begin{abstract}
This work introduces Knowledge-Distilled Physics-Informed Neural Networks (KD-PINN), a framework that transfers the predictive accuracy of a high-capacity teacher model to a compact student through a continuous adaptation of the Kullback-Leibler divergence. In order to confirm its generality for various dynamics and dimensionalities, the framework is evaluated on a representative set of partial differential equations (PDEs). Across the considered benchmarks, the student model achieves inference speedups ranging from ×4.8 (Navier–Stokes) to ×6.9 (Burgers), while preserving accuracy.
Accuracy is improved by on the order of $1\%$ when the model is properly tuned. The distillation process also revealed a regularizing effect. With an average inference latency of 5.3 ms on CPU, the distilled models enter the ultra-low-latency real-time regime defined by sub-10 ms performance. Finally, this study examines how knowledge distillation reduces inference latency in PINNs, to contribute to the development of accurate ultra-low-latency neural PDE solvers.
\end{abstract}

\bigskip


\section{Introduction}

Partial differential equations (PDEs) are mathematical representations that model a wide range of phenomena in physics and engineering. Common real-world applications surround us, such as heat and wave propagation, simulation and control, and financial forecasting.
The recent growing role of real-time simulation and control in both scientific and industrial contexts has increased the demand for fast and reliable PDE solvers. Among recent neural approaches, Physics-Informed Neural Networks (PINNs) \citep{Raissi2019} have provided a versatile method to remedy the limitations of early data-driven methods, that required a large amount of data.

PINNs incorporate the equations of physics governing the system as soft constraints during training, which ensures the physical consistency of the solutions even when data is scarce \citep{Raissi2019, Karniadakis2021}. Furthermore, they constitute a mesh-free approach, allowing them to adapt to complex geometries and high-dimensional domains, a situation not feasible with traditional numerical solvers \citep{Cuomo2022SciAdv}. Despite this conceptual shift, standard PINNs present optimization difficulties and slow convergence, as well as high computational training costs, which are mainly attributable to competing physics and data losses \citep{Wang2021Understand, Mishra2023, Lawal2022Review}. The inference phase is also computationally expensive due to the size of the network and nonlinear activations. This limitation, which we propose to overcome, is the one that restricts their use in real-time \citep{Wang2020PINNFail}.

Recent advances have mainly improved training stability or convergence efficiency \citep{Jagtap2020, McClenny2020SelfAdaptive, Psaros2022_MetaPINN}, whereas the inference phase has received far less attention.
In contrast, the proposed knowledge-distilled physics-informed network (KD-PINN) framework focuses on explicit inference-time acceleration while preserving physical consistency.
Recent distillation-inspired variants—such as SKD-PINN \citep{SKDPIN2025}, ID-PINN \citep{IDPINN2024}, and PINNCoM \citep{PINNCoM2025}—mainly aim to reduce model size or memory usage to improve compactness or training stability, but do not explicitly quantify latency or optimize execution speed. Reducing the memory footprint alone does not guarantee faster inference, since on modern GPUs the main limiting factors are network depth and the degree of kernel parallelism \citep{Han2015, Li2021_FNO}. 

KD-PINN transfers the predictive ability of a high-capacity teacher PINN to a compact student PINN model through a distillation process guided by both data and physics.
While classical distillation \citep{Hinton2015_KD} relies on probability distributions in classification tasks, its adaptation to regression replaces soft-label matching with the minimization of a continuous discrepancy between teacher and student predictions.  Under Gaussian and homoscedastic assumptions, this discrepancy can be expressed as a continuous Kullback–Leibler divergence:
\begin{equation}
\mathcal{L}_{\mathrm{KL}}
= \int_{\Omega} p_T(u)\,\log\!\frac{p_T(u)}{p_S(u)}\,du
\;\approx\;
\frac{1}{2}\int_{\Omega}\!\bigl|p_T(u)-p_S(u)\bigr|^2\,du,
\end{equation}
where \(p_T(u)\) and \(p_S(u)\) denote the normalized teacher and student predictive densities over the solution domain (see Section~\ref{subsec:kd-regression}). This formulation is proportional to the Mean Squared Error (MSE) which enables the distillation of a PINN model.
The distilled model with reduced sequential depth and lower arithmetic intensity, performs 5--10$\times$ faster inference than the teacher that is a standard PINN on a conventional CPU setup, without compromising accuracy.
In addition to speed improvements, the distillation process acts as an implicit regularizer that flattens the loss landscape and reduces gradient variances. Consequently, it reduces training instability and overfitting, and improves generalization. This effect is relevant to maintain inference accuracy in a student network that is deployed under limited computational resources, for example in CPU-based or embedded edge deployments.
The framework combines physical consistency with inference acceleration, and demonstrates sub-10 ms inference across multiple PDEs.
To the best of our knowledge, this constitutes the first PINN-based framework with quantified latency improvements validated under realistic deployment conditions, marking a concrete step from real-time feasibility to ultra-low-latency real-time capability (end-to-end responses below 10 ms) \citep{3gpp, Nasrallah2019ULL, Aijaz2015Tactile, Liang2022URLLC}. 

The remainder of this paper is organized as follows.
Section~\ref{sec:methods} presents the KD--PINN framework and its continuous distillation formulation, section~\ref{sec:results} reports cross-PDE evaluations and latency benchmarks, finally, section~\ref{sec:analysis-ratio} presents the theoretical limits of achievable speed-up.

\section{Method}
\label{sec:methods}

The methodological foundation of KD-PINN is to apply knowledge distillation to accelerate a PINN. The procedure involves two stages: first, training a large-scale PINN to solve the PDE target accurately (teacher), then transferring its learned representations to a reduced PINN model (student).

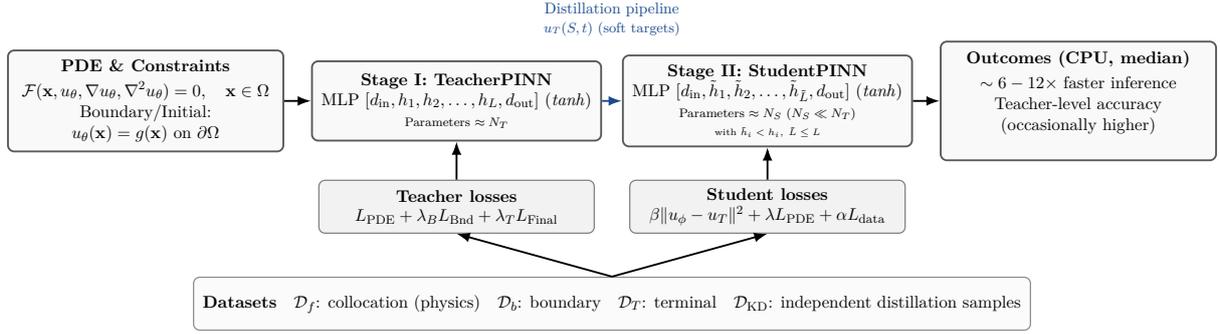
\begin{figure*}[t]
\centering
\captionsetup{justification=centering}
\resizebox{\linewidth}{!}{%
\begin{tikzpicture}[
  x=1cm, y=1cm, font=\normalsize, transform shape,
  box/.style   ={rounded corners, draw=black!65, very thick, fill=gray!5,
                 align=center, minimum width=6.2cm, minimum height=1.8cm, inner sep=6pt},
  sbox/.style  ={rounded corners, draw=black!50, thick, fill=gray!10,
                 align=center, minimum width=6.2cm, minimum height=1.2cm, inner sep=5pt},
  bar/.style   ={rounded corners, draw=black!45, thick, fill=gray!3,
                 align=left, minimum width=14.6cm, minimum height=1.2cm, inner sep=6pt},
  arrow/.style ={-{Latex[length=3mm]}, very thick},
  kdarrow/.style={-{Latex[length=3mm]}, very thick, draw=joBlue},
]

\coordinate (P) at (0,0);
\coordinate (T) at (7,0);
\coordinate (S) at (14,0);
\coordinate (R) at (21,0);

\coordinate (TL) at (7,-2.4);
\coordinate (SL) at (14,-2.4);
\coordinate (DB) at (10.5,-4.6);

\node[box] at (P)  (pde)     {\textbf{PDE \& Constraints}\\[2pt]
$\mathcal{F}(\mathbf{x}, u_\theta, \nabla u_\theta, \nabla^2 u_\theta)=0,\quad \mathbf{x}\in\Omega$\\
Boundary/Initial:\\ $u_\theta(\mathbf{x})=g(\mathbf{x})\ \text{on}\ \partial\Omega$};

\node[box] at (T)  (teacher) {\textbf{Stage I: TeacherPINN}\\[-1pt]
MLP $[d_{\text{in}}, h_1, h_2, \dots, h_L, d_{\text{out}}]$ (\textit{tanh})\\
\scriptsize Parameters $\approx N_T$};

\node[box] at (S)  (student) {\textbf{Stage II: StudentPINN}\\[-1pt]
MLP $[d_{\text{in}}, \tilde{h}_1, \tilde{h}_2, \dots, \tilde{h}_{\tilde{L}}, d_{\text{out}}]$ (\textit{tanh})\\
\scriptsize Parameters $\approx N_S$ ($N_S \ll N_T$)\\[-2pt]
{\tiny with $\tilde{h}_i < h_i,\;\tilde{L} \le L$}};

\node[box] at (R)  (results) {\textbf{Outcomes (CPU, median)}\\[2pt]
$\sim 6-12\times$ faster inference\\
Teacher-level accuracy\\ (occasionally higher)\\
};

\node[sbox] at (TL) (tloss) {\textbf{Teacher losses}\\[-1pt]
$L_{\mathrm{PDE}}+\lambda_B L_{\mathrm{Bnd}}+\lambda_T L_{\mathrm{Final}}$};

\node[sbox] at (SL) (sloss) {\textbf{Student losses}\\[-1pt]
$\beta\|u_\phi-u_T\|^2+\lambda L_{\mathrm{PDE}}+\alpha L_{\mathrm{data}}$};

\node[bar]  at (DB) (datasets) {\textbf{Datasets}\quad
$\mathcal{D}_f$: collocation (physics)\quad
$\mathcal{D}_b$: boundary\quad
$\mathcal{D}_T$: terminal\quad
$\mathcal{D}_{\mathrm{KD}}$: independent distillation samples};

\draw[arrow]   (pde.east)    -- (teacher.west);
\draw[kdarrow]
  (teacher.east) --
  node[midway, above=50pt, text=joBlue]{\small Distillation pipeline}
  node[midway, above=38pt,  text=joBlue]{\footnotesize $u_T(S,t)$ (soft targets)}
  (student.west);
\draw[arrow]   (student.east) -- (results.west);

\draw[arrow]   (tloss.north) -- (teacher.south);
\draw[arrow]   (sloss.north) -- (student.south);

\draw[arrow]   (datasets.north) -- (tloss.south);
\draw[arrow]   (datasets.north) -- (sloss.south);

\end{tikzpicture}%
}
\caption{Methodological overview of the proposed KD--PINN framework.}
\label{fig:method_overview}
\end{figure*}

\subsection{Stage I - Training a high-fidelity PINN}
\label{ssec:pinn}

Let $\mathcal{N}[u](S,t)=0$ be a PDE on
$\Omega=[S_{\min},S_{\max}]\times[0,T]$ with solution $u(S,t)$.
TeacherPINN is a PINN $u_{\theta}(S,t)$ \citep{Raissi2019}, parametrised by
$\theta = \{W^{(l)}, b^{(l)}, \sigma\}_{l=1}^{L}$, where $W^{(l)} \in \mathbb{R}^{d_{l} \times d_{l-1}}$, $b^{(l)} \in \mathbb{R}^{d_{l}}$, and $\sigma$ is a nonlinear activation function, that minimises the composite loss
\begin{align}
\mathcal{L}_{\text{PINN}}(\theta)=
\underbrace{\frac{1}{N_{u}}\sum_{i=1}^{N_{u}}
\lvert u_{\theta}(S_{u}^{i},t_{u}^{i})-u^{\text{obs}}(S_{u}^{i},t_{u}^{i})\rvert^{2}}
_{\mathcal{L}_{\text{data}}}
+
\underbrace{\frac{1}{N_{f}}\sum_{j=1}^{N_{f}}
\lvert\mathcal{N}[u_{\theta}](S_{f}^{j},t_{f}^{j})\rvert^{2}}
_{\mathcal{L}_{\text{PDE}}}.
\label{eq:pinn-loss}
\end{align}
Training datasets are $\mathcal{D}_u = \{(S_u^i, t_u^i, u^{\text{obs}}(S_u^i, t_u^i))\}_{i=1}^{N_u}$ with boundary and terminal values and $\mathcal{D}_f = \{(S_f^j, t_f^j)\}_{j=1}^{N_f}$ with points to enforce the PDE.
These points can either coincide or be sampled independently. Using distinct sets allows the data constraints and the physics residual to be enforced at different locations, which reduces competition between the two losses and enables targeted sampling of the residual where it is most informative. This separation becomes particularly beneficial when combined with adaptive weighting or residual-based sampling strategies \citep{subramanian2022adaptive, Wang2021Understand}.

The training objective consists in finding the parameters minimizing the total loss,
\(
\theta^\star = \arg\min_\theta\,\mathcal{L}_{\text{PINN}}(\theta),
\)
which is iteratively approached by stochastic gradient descent as
\begin{equation}
\theta^{(k+1)} = \theta^{(k)} - \eta\,\nabla_\theta \mathcal{L}_{\text{PINN}}(\theta^{(k)}),
\end{equation}
where $\eta$ denotes the learning rate.
In practice, the Adam optimizer is employed for rapid exploration of the parameter space,
followed by a quasi--Newton refinement with the L--BFGS algorithm to ensure smooth convergence toward a stable minimum.

\paragraph{Black–Scholes specialisation}
For European options,
\begin{equation}
u_t + rSu_S + \tfrac{1}{2}\sigma^2 S^2u_{SS} - ru = 0,\qquad
g(S)=u(S,T)=\max(S-K,0).
\label{eq:bs-pde}
\end{equation}
We impose boundary values at $S_{\min}$, $S_{\max}$ and the terminal payoff condition. They are enforced through the following :
\begin{equation}
\label{eq:bs-loss}
\mathcal{L}_{\text{Bnd}} = \frac{1}{N_b} \sum_{i=1}^{N_b} \left| u_\theta(S_b^i, t_b^i) - u^{\text{obs}}(S_b^i, t_b^i) \right|^2, \quad
\mathcal{L}_{\text{Final}} = \frac{1}{N_T} \sum_{i=1}^{N_T} \left| u_\theta(S_T^i, T) - g(S_T^i) \right|^2
\end{equation}
where $N_b$ and $N_T$ denote the number of points sampled at the spatial boundary and final time, respectively.
The total loss becomes:

\begin{align}
\mathcal{L}_{\text{PINN}}(\theta) =\ 
& \frac{1}{N_f} \sum_{j=1}^{N_f} \left| \mathcal{N}[u_\theta](S_f^j, t_f^j) \right|^2 \nonumber \\
& + \lambda \left(
    \frac{1}{N_b} \sum_{i=1}^{N_b} \left| u_\theta(S_b^i, t_b^i) - u^{\text{obs}}(S_b^i, t_b^i) \right|^2
    + \frac{1}{N_T} \sum_{i=1}^{N_T} \left| u_\theta(S_T^i, T) - \max(S_T^i - K, 0) \right|^2
\right)
\end{align}

\subsection{Stage II - Distilling speed into a StudentPINN}
\label{ssec:kd-pinn}

\subsubsection{Knowledge distillation: from cross-entropy to Kullback-Leibler divergence}

Knowledge Distillation (KD) is a compression technique introduced by Hinton et al. \citep{Hinton2015_KD}, where a compact student model mimics a larger teacher model. The student learns from both ground truth labels (hard targets) and teacher predictions (soft targets). To achieve this, according to Hinton \textit{et al.}, knowledge distillation is most effectively carried out by minimising the cross-entropy between the teacher’s softened outputs and the student’s predictions. Consequently, knowledge distillation can also be interpreted as the minimization of the Kullback-Leibler divergence \citep{kullback_information_1951} between the teacher’s softened outputs and the student’s predictions \citep{CoverThomas2006},
\begin{equation}\mathcal{L}_{\text{KL}} = D_{\text{KL}}(\boldsymbol{q} \| \boldsymbol{p}) = \sum_{i=1}^C q_i \log \left( \frac{q_i}{p_i} \right),\end{equation} where C is the number of classes. 
\paragraph{Derivation} 
Assuming an input sample \( x \in \mathcal{X} \) with ground-truth label \( y \in \{1, \dots, C\} \), 
the student and teacher models, parameterized by \( f_S(\cdot) \) and \( f_T(\cdot) \), 
produce class-probability distributions $p(x) = softmax(f_S(x)) \in [0,1]^C$ and $q(x) = softmax(f_T(x)/T) \in [0,1]^C$
where \( T > 1 \) is the temperature used to soften the teacher’s output distribution.\\
Since \(\text{CE}(\boldsymbol{q}(x), \boldsymbol{p}(x)) 
= - \sum_{i=1}^{C} q_i(x) \log p_i(x),
\label{eq:ce-onehot}
\)
and \( H(\boldsymbol{q}(x)) = - \sum_{i=1}^{C} q_i(x) \log q_i(x) \),
\begin{equation}
\mathcal{L}_{\text{soft}} = \text{CE}(\boldsymbol{q}(x), \boldsymbol{p}(x)) 
= - \sum_{i=1}^{C} q_i(x) \log p_i(x)
= D_{\text{KL}}(\boldsymbol{q}(x) \| \boldsymbol{p}(x)) + H(\boldsymbol{q}(x)),
\end{equation}
As $H(\mathbf{q})$ is independent of the student, minimizing \( \mathcal{L}_{\text{soft}} \) equals minimizing KL divergence
\begin{equation}
\arg\min_{\mathbf{p}} CE(\mathbf{q},\mathbf{p})
\;\equiv\; \arg\min_{\mathbf{p}} D_{\text{KL}}(\mathbf{q} \| \mathbf{p}).
\label{eq:ce-vs-kl}
\end{equation}
\(\mathcal{L}_{\text{hard}} = \text{CE}(y, \boldsymbol{p}(x)) = -\log p_y(x)\), and \(p_y(x)\) is the probability assigned by the model to the true class \(y\), and the expression of the loss is \begin{equation}\mathcal{L}_{\text{KD-total}} = \lambda_{\text{hard}} \cdot \mathcal{L}_{\text{CE}} + \lambda_{\text{soft}} \cdot \mathcal{L}_{\text{KL}},\end{equation}

While both formulations are equivalent in the discrete (classification) setting, we adopt the Kullback–Leibler divergence,
as this choice facilitates the extension to the continuous regime of PINNs under Gaussian assumptions (see Section~\ref{subsec:kd-regression}).

\subsubsection{From classification to regression.}
\label{subsec:kd-regression}

In regression settings such as PINNs, outputs are continuous scalars $u(S,t) \in \mathbb{R}$, the softmax-based KL divergence is then inapplicable. However, under a Gaussian approximation, the KL divergence between two univariate normal distributions with equal variance reduces to the squared distance between their means \citep{Bishop2006}. In fact, the outputs of teacher and student models can be interpreted probabilistically \citep{ahn_variational_2019} rather than deterministically: a PINN provides pointwise predictions $u(S,t)$ that can be viewed as realizations of random variables affected by model uncertainty, numerical error, and stochastic training dynamics \citep{ahn_variational_2019, Linka2022BayesianPINN}. 
Accordingly, at any point $(S,t)$, the teacher and student predictions can be modeled respectively as
\[
u_T(S,t) \sim \mathcal{N}(\mu_T, \sigma_T^2),
\qquad
u_{\phi}(S,t) \sim \mathcal{N}(\mu_{\phi}, \sigma_{\phi}^2),
\]
where the mean represents the deterministic network output and the variance encodes local epistemic and numerical uncertainty. 
The Gaussian assumption is justified by the central limit behavior of residual errors. The closed-form expression of the Kullback--Leibler divergence between two Gaussian distributions can be derived analytically ~\citep{CoverThomas2006, Bishop2006} as
\begin{equation}
D_{\text{KL}}\!\left(
\mathcal{N}(\mu_T, \sigma_T^2) 
\;\middle\|\;
\mathcal{N}(\mu_{\phi}, \sigma_{\phi}^2)
\right)
= 
\frac{1}{2}\!\left[
\frac{(\mu_T - \mu_{\phi})^2}{\sigma_{\phi}^2}
+ 
\frac{\sigma_T^2}{\sigma_{\phi}^2}
- 1 
- \ln\!\frac{\sigma_T^2}{\sigma_{\phi}^2}
\right].
\label{eq:kl-general}
\end{equation}
Assuming homoscedasticity ($\sigma_T^2 = \sigma_\phi^2 = \sigma^2$) yields
\begin{equation}
D_{\text{KL}}\!\left(
\mathcal{N}(\mu_T, \sigma^2)
\;\middle\|\;
\mathcal{N}(\mu_{\phi}, \sigma^2)
\right)
= 
\frac{|\mu_T - \mu_{\phi}|^2}{2\sigma^2}
\propto 
|u_T - u_{\phi}|^2
\label{eq:kl-to-mse}
\end{equation}
Hence, the Kullback–Leibler divergence reduces to a scaled mean-squared deviation between the teacher and student predictions, which justifies the use of the MSE form of the distillation loss in continuous regression regimes.
Thus, at any point $\mathbf{x} = (S, t)$, the soft distillation objective becomes
\begin{equation}
\mathcal{L}_{\text{soft}}=\mathcal{L}_{\text{KD(reg)}}(\mathbf{x}) =
|u_{\phi}(\mathbf{x}) - u_T(\mathbf{x})|^2.
\label{eq:kd-mse}
\end{equation}
Then,
\begin{equation}
\mathcal{L}_{\text{KD}}(\phi) = \frac{1}{N_k} \sum_{k=1}^{N_k} \left| u_\phi(S_k, t_k) - u_T(S_k, t_k) \right|^2.
\end{equation}
The hard loss is defined as the squared residual of the governing PDE, that acts as the hard label in classification (ground truth)
\begin{equation}
\mathcal{L}_{\text{hard(reg)}}(\mathbf{x}) =
|\mathcal{N}[u_{\phi}](\mathbf{x})|^2 =\mathcal{L}_{\text{PDE}}.
\label{eq:kd-hard}
\end{equation}

\noindent Hence,
\begin{equation}
\begin{aligned}
\mathcal{L}_{\text{KD-PINN}}(\phi) &=
\lambda_{\mathrm{PDE}}
\underbrace{\frac{1}{N_f}\!\sum_{j=1}^{N_f}
\!\left| \mathcal{N}[u_{\phi}](S_f^j,t_f^j)\right|^2}_{\mathcal{L}_{\mathrm{PDE}}^{(S)}}
+ \lambda_{\mathrm{data}}
\underbrace{\frac{1}{N_d}\!\sum_{i=1}^{N_d}
\!\left| u_{\phi}(S_d^i,t_d^i) - u^{\mathrm{obs}}(S_d^i,t_d^i)\right|^2}_{\mathcal{L}_{\mathrm{data}}^{(S)}}
\\[3pt]
&\quad
+ \lambda_{\mathrm{KD}}
\underbrace{\frac{1}{N_k}\!\sum_{k=1}^{N_k}
\!\left| u_{\phi}(S_k,t_k) - u_T(S_k,t_k)\right|^2}_{\mathcal{L}_{\mathrm{KD}}^{(S)}}.
\end{aligned}
\label{eq:total-kd-pinn-condensed}
\end{equation}

\subsubsection{Application to Black-Scholes}

TeacherPINN $u_T(S,t)$ is trained using the composite loss
\begin{equation}
\mathcal{L}_{\text{Teacher}}(\theta_T) 
= \mathcal{L}_{\text{PDE}}^{(T)} 
+ \lambda_{\text{B}}\,\mathcal{L}_{\text{Bnd}}^{(T)} 
+ \lambda_{\text{T}}\,\mathcal{L}_{\text{Final}}^{(T)},
\label{eq:teacher-loss}
\end{equation}
where the individual terms are defined as\\
$\mathcal{L}_{\text{PDE}}^{(T)} = \tfrac{1}{N_f} \sum_{j=1}^{N_f} |\mathcal{N}[u_T](S_f^j, t_f^j)|^2$,  
$\mathcal{L}_{\text{Bnd}}^{(T)} = \tfrac{1}{N_b} \sum_{i=1}^{N_b} |u_T(S_b^i, t_b^i) - u^{\text{obs}}(S_b^i, t_b^i)|^2$,  
and\\  
$\mathcal{L}_{\text{Final}}^{(T)} = \tfrac{1}{N_T} \sum_{i=1}^{N_T} |u_T(S_T^i, T) - g(S_T^i)|^2$.\\
\noindent with \quad
$\mathcal{D}_f=\{(S_f^{\,j},t_f^{\,j})\}_{j=1}^{N_f}$,\;
$\mathcal{D}_b=\{(S_b^{\,i},t_b^{\,i},u^{\mathrm{obs}})\}_{i=1}^{N_b}$,\;
$\mathcal{D}_T=\{(S_T^{\,i},T)\}_{i=1}^{N_T}$.
Here, $\mathcal{D}_f$ denotes the set of $N_f$ collocation points, 
$\mathcal{D}_b$ the $N_b$ boundary samples, 
and $\mathcal{D}_T$ the $N_T$ terminal points.

StudentPINN $u_\phi(S,t)$, is then trained on the same datasets $\mathcal{D}_f$, $\mathcal{D}_b$, and $\mathcal{D}_T$, while also incorporating soft supervision from the teacher via a distillation set $\mathcal{D}_{\text{KD}} = \{(S_k, t_k)\}_{k=1}^{N_k}$, sampled independently in the same domain without coinciding with $\mathcal{D}_f$, $\mathcal{D}_b$, or $\mathcal{D}_T$. This independent sampling ensures that the teacher provides complementary guidance on additional points outside of the datasets used for physics and boundaries. 
The KD--PINN loss combines physical, boundary, terminal, 
and distillation terms as
\begin{equation}
\mathcal{L}_{\mathrm{KD\text{-}PINN}}(\phi)
= \lambda_{\mathrm{PDE}}\mathcal{L}_{\mathrm{PDE}}^{(S)}
+ \lambda_{\mathrm{BC}}\mathcal{L}_{\mathrm{BC}}^{(S)}
+ \lambda_{\mathrm{T}}\mathcal{L}_{\mathrm{T}}^{(S)}
+ \lambda_{\mathrm{KD}}\mathcal{L}_{\mathrm{KD}}^{(S)},
\label{eq:kd-pinn-bs-total}
\end{equation}
where $\mathcal{L}_{\mathrm{PDE}}^{(S)}$, 
$\mathcal{L}_{\mathrm{BC}}^{(S)}$, 
and $\mathcal{L}_{\mathrm{T}}^{(S)}$
retain the same formulation as in the TeacherPINN objective 
(see Eq.~\eqref{eq:teacher-loss}),
and the additional distillation term is defined as
\begin{equation}
\mathcal{L}_{\mathrm{KD}}^{(S)}
= \frac{1}{N_k}\sum_{k=1}^{N_k}
|u_\phi(S_k,t_k)-u_T(S_k,t_k)|^2,
\end{equation}
defined over the distillation dataset 
$\mathcal{D}_{\mathrm{KD}}$.

\subsection{Implementation, evaluation, and convergence analysis}
\label{ssec:implementation-evaluation}

\subsubsection{Protocol and implementation}

\begin{algorithm}[t]
\caption{KD--PINN training for Black--Scholes}
\label{alg:kdpinn}
\begin{algorithmic}[1]
\Require Black--Scholes params $(K,r,\sigma,T)$; domains $\Omega_{\mathrm{int}},\Gamma_{\mathrm{bc}},\Gamma_{\mathrm{term}}$; weights $(w_{\mathrm{PDE}},w_{\mathrm{BC}},w_{\mathrm{term}},w_{\mathrm{KD}})$; temperature $\tau$; batch sizes $(n_{\mathrm{int}},n_{\mathrm{bc}},n_{\mathrm{term}},n)$; epochs $(N_T,N_S)$; optimizer hyperparams
\Ensure Trained student network $u_{\phi}$
\State Initialize teacher $u_{\theta}$ and student $u_{\phi}$
\Statex \textbf{Teacher pretraining}
\For{$e=1$ to $N_T$}
  \State Sample $\{(S_i,t_i)\}_{i=1}^{n_{\mathrm{int}}}\!\sim\!\Omega_{\mathrm{int}}$, $\{(S_j,t_j)\}_{j=1}^{n_{\mathrm{bc}}}\!\sim\!\Gamma_{\mathrm{bc}}$, $\{(S_k,T)\}_{k=1}^{n_{\mathrm{term}}}\!\sim\!\Gamma_{\mathrm{term}}$
  \State Compute physics \& constraints:
  \Statex \hspace{\algorithmicindent}$L_{\mathrm{PDE}}(\theta),\, L_{\mathrm{BC}}(\theta),\, L_{\mathrm{term}}(\theta)$
  \State Update $\theta \gets \mathrm{OptStep}\!\left(\theta,\, w_{\mathrm{PDE}}L_{\mathrm{PDE}} + w_{\mathrm{BC}}L_{\mathrm{BC}} + w_{\mathrm{term}}L_{\mathrm{term}}\right)$
\EndFor
\Statex \textbf{Student distillation}
\For{$e=1$ to $N_S$}
  \State Sample minibatches as above; form teacher targets $u_T(S,t)\gets u_{\theta}(S,t)$
  \State Compute physics/constraints for $u_{\phi}$:
  \Statex \hspace{\algorithmicindent}$L_{\mathrm{PDE}}(\phi),\, L_{\mathrm{BC}}(\phi),\, L_{\mathrm{term}}(\phi)$
  \State Compute distillation loss (e.g., MSE or Huber):
  \Statex \hspace{\algorithmicindent}$L_{\mathrm{KD}}(\phi)\gets \frac{1}{n}\!\sum \bigl|u_{\phi}(S,t) - u_T(S,t)\bigr|^{2}$
  \State Update $\phi \gets \mathrm{OptStep}\!\left(\phi,\, w_{\mathrm{PDE}}L_{\mathrm{PDE}} + w_{\mathrm{BC}}L_{\mathrm{BC}} + w_{\mathrm{term}}L_{\mathrm{term}} + w_{\mathrm{KD}}L_{\mathrm{KD}}\right)$
  \State (Optional) curriculum/informed sampling; early stopping on validation
\EndFor
\State \Return $u_{\phi}$
\end{algorithmic}
\end{algorithm}

\paragraph{Evaluation}
\label{par:evaluation-protocol}
StudentPINN is compared to TeacherPINN under identical settings, and both models are trained independently on non-overlapping points for collocations, boundaries, and terminal conditions. Then they are evaluated on a uniform grid covering $(S,t)$ with the analytical Black--Scholes solution $u^{*}(S,t)$ as reference.
Predictive accuracy is quantified by the root-mean-square error (RMSE)
\begin{equation}
\mathrm{RMSE}
= \sqrt{\tfrac{1}{N}\sum_{i=1}^{N}
\!\bigl(u_{\phi}(S_i,t_i)-u^{*}(S_i,t_i)\bigr)^{2}},
\label{eq:rmse}
\end{equation}
and by the relative $\mathcal{L}_2$ error
\begin{equation}
\mathrm{relL2}
= \tfrac{\|u_{\phi}-u^{*}\|_{2}}{\|u^{*}\|_{2}}.
\label{eq:relL2}
\end{equation}

Computational efficiency is assessed by the median inference latency per forward pass, measured on identical CPU
and GPU configurations.

\subsubsection{Effect of architecture on PINN accuracy}\label{ssec: arch-proof} 

\paragraph{Justification of experimental choices}
From a representational perspective, when the architecture is enlarged, the set of functions the network is able to represent expands. A larger or deeper model therefore has, in principle, access to a lower approximation error and may capture finer details of the solution. This theoretical argument does not guarantee a smaller training error in practice, as PINNs often face optimization difficulties. Nevertheless, a high-capacity teacher provides a more accurate target for distillation.
In the experiments, architectural configurations were selected in accordance with guidelines reported in the literature. They range from 4 to 8 hidden layers with 50 to 200 neurons per layer \citep{luo_physics-informed_2025, harmening_effect_2024}, in function of the complexity and stiffness of the PDE.

\subsubsection{Convergence analysis}

Following Raissi~\emph{et al.}~\citep{Raissi2019},
Shin~\emph{et al.}~\citep{Shin2020},
Wu~\emph{et al.}~\citep{Wu2023_ConvergencePINN}, and
Mishra \& Molinaro~\citep{Mishra2023},
the physics loss in PINNs can be interpreted as an empirical
$\mathcal{L}^{2}$-norm of the residual operator
$\mathcal{F}(u)\vcentcolon=\mathcal{N}[u]$.
This section extends these convergence and stability analyses
to the knowledge-distilled setting, where a student model
$u_{\phi}$ learns from a high-fidelity teacher~$u_{T}$.

\begin{theorem}[Residual and error transfer under KD]
\label{thm:kdpinn-transfer}
Let $u^\star$ be the exact solution of $\mathcal{F}(u)=0$ on a bounded domain $\Omega$, and let $u_T$ (teacher) and $u_\phi$ (student) be two candidate solutions.
Assume:
\begin{enumerate}[label=(\roman*),nosep,leftmargin=2.2em]
\item \textbf{Teacher accuracy:} $\|u_T-u^\star\|_{L^2(\Omega)} \le \varepsilon_T$;
\item \textbf{Teacher consistency:} $\|\mathcal{F}(u_T)\|_{L^2(\Omega)} \le \delta_T$;
\item \textbf{Student--teacher proximity:} $\|u_\phi-u_T\|_{L^2(\Omega)} \le \varepsilon_D$;
\item \textbf{Lipschitz residual:} $\|\mathcal{F}(u)-\mathcal{F}(v)\|_{L^2(\Omega)} \le L\,\|u-v\|_{L^2(\Omega)}$ for all $u,v$.
\end{enumerate}
Then
\begin{equation}\label{eq:residual-bound}
\|\mathcal{F}(u_\phi)\|_{L^2(\Omega)} \;\le\; L\,\varepsilon_D + \delta_T.
\end{equation}
If, in addition, the PDE satisfies a stability estimate (Hadamard well-posedness) 
\begin{equation}\label{eq:stability}
\|v-u^\star\|_{L^2(\Omega)} \;\le\; \kappa\,\|\mathcal{F}(v)\|_{L^2(\Omega)}
\quad\text{for all }v\text{ near }u^\star,
\end{equation}
then the student inherits the non-asymptotic bound
\begin{equation}\label{eq:error-bound}
\|u_\phi-u^\star\|_{L^2(\Omega)} \;\le\;
\varepsilon_T \;+\; \kappa\,(L\,\varepsilon_D+\delta_T).
\end{equation}
\end{theorem}

\emph{Proof.}
By the triangle inequality and assumption~(ii), we have
\[
\|\mathcal{F}(u_\phi)\|_{L^2}
  \le \|\mathcal{F}(u_\phi)-\mathcal{F}(u_T)\|_{L^2}
     +\|\mathcal{F}(u_T)\|_{L^2}
  \le L\,\varepsilon_D + \delta_T,
\]
which yields~\eqref{eq:residual-bound}.
Combining this bound with the stability estimate~\eqref{eq:stability}
and the teacher accuracy~(i) leads to
\eqref{eq:error-bound}.
\hfill$\square$

\paragraph{Justification of (iv)}
For linear PDEs with bounded coefficients
(e.g.\ Black--Scholes), the residual operator is bounded and therefore
Lipschitz on compact domains~\citep{evans_partial_2010}.
For semilinear or mildly nonlinear PDEs,
this holds whenever the nonlinear term is Lipschitz in~$u$
\citep{Wu2023_ConvergencePINN,Mishra2023}.

\section{Experiments and Results}
\label{sec:results}

\subsection{In-domain experiment on the Black-Scholes equation}

\subsubsection{Experimental setup}

\paragraph{Training configuration}
The benchmark is the Black--Scholes PDE for European options~\eqref{eq:bs-pde} with $(K,r,\sigma,T)=(1,0.05,0.2,1)$, trained over $(S,t)\!\in[0.5,1.5]\times[0,1]$ using the boundary and terminal losses~\eqref{eq:bs-loss}. 
TeacherPINN and StudentPINN adopt the standard fully-connected architecture~\citep{Raissi2019}, with layers $[2,50,50,50,1]$ and $[2,20,20,20,1]$, respectively, \texttt{tanh} activations, linear output, and Xavier initialization. 
In this configuration, the number of PINN parameters decreases by almost an order of magnitude, from approximately $8{,}051$ to $1{,}061$. 
Inputs $(S,t)$ are linearly rescaled to $[-1,1]^2$ to preserve the effective range of $\tanh$, which ensures smooth derivatives $u$, $u_{S}$, $u_{SS}$ for stable automatic differentiation.
The teacher minimizes $\mathcal{L}_T = w_{\mathrm{PDE}}\mathcal{L}_{\mathrm{PDE}} + w_{\mathrm{BC}}\mathcal{L}_{\mathrm{BC}} + w_{\mathrm{Term}}\mathcal{L}_{\mathrm{Term}}$
with weights $(1,12,15)$, while boundary and terminal losses use a Huber penalty~\citep{Huber1964} for robustness near the payoff kink. 
The student minimizes $\mathcal{L}_S = \mathcal{L}_T + w_{\mathrm{KD}}\mathcal{L}_{\mathrm{KD}}$ with $\mathcal{L}_{\mathrm{KD}}=\mathrm{MSE}(u_S/\tau,u_T/\tau)$, $w_{\mathrm{KD}}=1.5$, and $\tau=1.25$.
Collocation points follow a 2D Sobol sequence with Owen scrambling~\citep{sobol1967,Owen1995}, providing uniform coverage, reduced variance, and faster convergence. 
Both networks use Adam ($\mathrm{lr}=10^{-3}$) for 8000 and 6000 iterations, respectively, followed by a 300-step fine-tuning phase at $10^{-4}$ to refine physical consistency. 
Each iteration samples 4096 collocation, 256 boundary, and 512 terminal points; the best checkpoint (lowest total loss) is retained for evaluation.

\paragraph{Evaluation protocol}
Accuracy and inference latency are evaluated after training. 
Accuracy is measured on a $100\times50$ uniform grid in $(S,t)$ by the comparison of $u_{\theta}(S,t)$ and the analytical Black--Scholes solution $u^{\star}(S,t)$, from which the RMSE and the relative $L^2$ error are calculated over $N=5000$ points. 
Latency is measured on CPU over $20{,}000$ uniformly sampled inputs $X_i=(S_i,t_i)$, rescaled to $[-1,1]^2$, under single-thread conditions (\texttt{OMP/MKL\_NUM\_THREADS}=1, \texttt{torch.set\_num\_threads(1)}). 
After 20 warm-up passes, 100 timed runs are averaged using \texttt{time.perf\_counter()}. The reported latency is the median ratio (Teacher/Student) in order to mitigate system noise.
All experiments run in double precision on an NVIDIA~T4 (16\,GB) and Intel~Xeon (2\,vCPUs, 13\,GB RAM) under Ubuntu~20.04 with PyTorch~2.2, NumPy~1.26, and SciPy~1.12. 
Logs, weights, and latency traces are archived for reproducibility.

\subsubsection{Performance and evaluation}
\label{subsec:In-domain accuracy and dynamic}

\begin{table}[H]
  \centering
  \caption{In-domain summary (Teacher vs Student).}
  \label{tab:T1}
  \begin{tabular}{l
                  S[table-format=1.4e-2]
                  S[table-format=1.4e-2]
                  S[table-format=3.2, parse-numbers = false]
                  l}
    \toprule
    {Model} & {RMSE} & {relL2} & {Latency (ms)} & {Ratio} \\
    \midrule
    Teacher & 2.2876131e-3 & 1.01809547e-2 & 48.87 & -- \\
    Student & 2.2857368e-3 & 1.01726043e-2 &  7.22 & {\(\times\)6.77} \\
    \bottomrule
  \end{tabular}
\end{table}

Table~\ref{tab:T1}, shows the efficiency gain induced by knowledge distillation. TeacherPINN exhibits a median latency of \SI{48.87}{\milli\second}, 
while StudentPINN requires only \SI{7.22}{\milli\second}, corresponding to a \(6.8\times\) speed-up. 
This advantage does not compromise accuracy: the Student even slightly improves performance, 
with RMSE decreasing from \num{2.2876e-3} to \num{2.2857e-3} and rel-$L^2$ from \num{1.01810e-2} to \num{1.01726e-2}, 
a relative improvement of 0.08\% across both metrics.

\subsubsection{Distillation dynamics}

\begin{figure*}[t]
  \centering
  \includegraphics[width=0.7\linewidth]{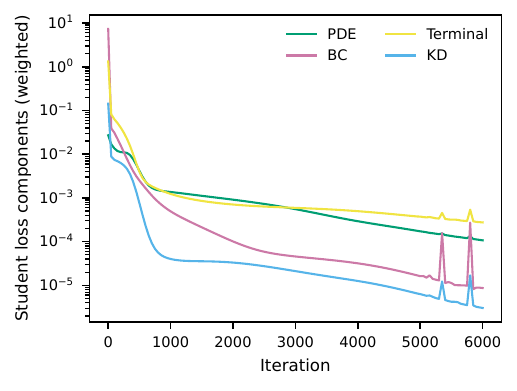}
  \caption{Loss decomposition for the StudentPINN}
  \label{fig:F1}
\end{figure*}

\paragraph{Loss decomposition}
To assess the training dynamics and the impact of knowledge distillation, we examine in Figure~\ref{fig:F1} the evolution of the components of the Student loss
\(
\mathcal{L}(\theta)
= w_{\mathrm{PDE}}\mathcal{L}_{\mathrm{PDE}}(\theta)
+ w_{\mathrm{BC}}\mathcal{L}_{\mathrm{BC}}(\theta)
+ w_{\mathrm{KD}}\mathcal{L}_{\mathrm{KD}}(\theta).
\label{eq:loss_total}
\)

It can be observed that The KD term, $\mathcal{L}_{\mathrm{KD}}(\theta)$, starts at a comparable level to $\mathcal{L}_{\mathrm{BC}}$ and $\mathcal{L}_{\mathrm{PDE}}$ (around $10^{0}$–$10^{-1}$) but decreases much faster—by more than three orders of magnitude within the first 1000 iterations—reaching below $10^{-4}$ near iteration~2000. 
This rapid decay originates from a dual dominance of the distillation term, both in amplitude 
and in direction. 
In fact, at initialization, the student's parameters $\theta$ are random, while the teacher 
already provides converged targets $u_T$ that satisfy the PDE and boundary conditions.
The distillation loss defined as
\(
\mathcal{L}_{\mathrm{KD}} = \|u_\theta - u_T\|^2,
\)
is a quadratic convex function of the network outputs.  

Its gradient
\(
\nabla_\theta \mathcal{L}_{\mathrm{KD}} = 2(u_\theta - u_T)\,\nabla_\theta u_\theta,
\)
is linear and proportional to the prediction error, which prevents strong output variations. 
In addition, at initialization, the student parameters are distant from those of the teacher, 
the gradient is then large in norm.

In contrast, the PDE and boundary losses,
\(
\mathcal{L}_{\mathrm{PDE}} = \|\mathcal{N}[u_\theta]\|^2,
\)
involve differential operators \((\partial_t, \partial_S, \partial_{SS})\).
Their resulting gradients include nested derivatives such as \(\nabla_\theta u_t\) 
and \(\nabla_\theta u_{SS}\),  
which amplify numerical noise and produce oscillatory updates across collocation points.
Formally, in gradient-based optimization, each update follows
\begin{equation}
\theta_{t+1} = \theta_t - \eta \nabla_\theta \mathcal{L}(\theta_t),
\qquad 
\nabla_\theta \mathcal{L} = \tfrac{1}{N}\sum_{i=1}^N \nabla_\theta \ell_i(\theta),
\end{equation}
so that the global gradient is the vector sum of all local gradients.  
For the PDE residual loss
\begin{equation}
\mathcal{L}_{\mathrm{PDE}} = \tfrac{1}{N}\sum_{i=1}^{N}r_i(\theta)^2,
\qquad
r_i = \mathcal{N}[u_\theta](x_i)-f(x_i),
\end{equation}
one obtains
\begin{equation}
\nabla_\theta \mathcal{L}_{\mathrm{PDE}}
=\tfrac{2}{N}\sum_{i=1}^N r_i(\theta)\,J_i(\theta),
\qquad
J_i=\nabla_\theta\mathcal{N}[u_\theta](x_i)=\nabla_\theta r_i,
\end{equation}
whose squared norm expands as
\begin{equation}
\|\nabla_\theta \mathcal{L}_{\mathrm{PDE}}\|^2
=\tfrac{4}{N^2}\sum_{i,j}r_i r_j\langle J_i,J_j\rangle.
\end{equation}
Each derivative ($\partial_S u_\theta$, $\partial_{SS}u_\theta$, etc.) changes sign with the local curvature of $u_\theta$: it is positive in convex regions and negative in concave ones, so that the residuals oscillate spatially as 
$r_i \propto \partial_t u_\theta + \partial_{SS}u_\theta + \cdots$.
Sign variations of $r_i$ and heterogeneous operators $J_i$ cause partial cancellations between cross-terms $(i\neq j)$ in $\nabla_\theta \mathcal{L}_{\mathrm{PDE}}$, which gives
\(
\|\nabla_\theta \mathcal{L}_{\mathrm{PDE}}\|^2
\ll \tfrac{4}{N^2}\Big(\sum_i |r_i|\|J_i\|\Big)^2.
\)
Hence, the effective gradient becomes weak in both norm and direction, and thus produces small steps
\(
\|\Delta\theta_t\|=\eta\|\nabla_\theta\mathcal{L}_{\mathrm{PDE}}\|
\)
and low directional coherence 
\(
\cos\!\big(\nabla_\theta\mathcal{L}_{\mathrm{PDE}}^{(t)},
           \nabla_\theta\mathcal{L}_{\mathrm{PDE}}^{(t-1)}\big)
=
\frac{
\left\langle
\nabla_\theta\mathcal{L}_{\mathrm{PDE}}^{(t)},
\nabla_\theta\mathcal{L}_{\mathrm{PDE}}^{(t-1)}
\right\rangle
}{
\|\nabla_\theta\mathcal{L}_{\mathrm{PDE}}^{(t)}\|\,
\|\nabla_\theta\mathcal{L}_{\mathrm{PDE}}^{(t-1)}\|
}
\!\approx\!0,
\)
which results in oscillatory updates and relative stagnation of learning.

As a result,
\(
\|\nabla_\theta \mathcal{L}_{\mathrm{KD}}\| \gg
\|\nabla_\theta \mathcal{L}_{\mathrm{PDE}}\|
\) and
\(\mathrm{Var}(\nabla_\theta \mathcal{L}_{\mathrm{KD}}) \ll
\mathrm{Var}(\nabla_\theta \mathcal{L}_{\mathrm{PDE}}),
\)
so the total update
\(
\Delta\theta = -\eta
\big(
\nabla_\theta \mathcal{L}_{\mathrm{KD}}
+ \nabla_\theta \mathcal{L}_{\mathrm{PDE}}
+ \nabla_\theta \mathcal{L}_{\mathrm{BC}}
\big)
\)
is driven mainly by the teacher–student loss, in a strong and consistent direction.  
The weight $w_{\mathrm{KD}}=1.5$ amplifies this effect, to ensure that distillation guides 
the optimization before the physics terms dominate.

\begin{figure*}[t]
  \centering
  \captionsetup[subfigure]{aboveskip=2pt,belowskip=1pt,justification=centering}

  \begin{subfigure}[b]{0.65\linewidth}
    \centering
    \includegraphics[width=\linewidth]{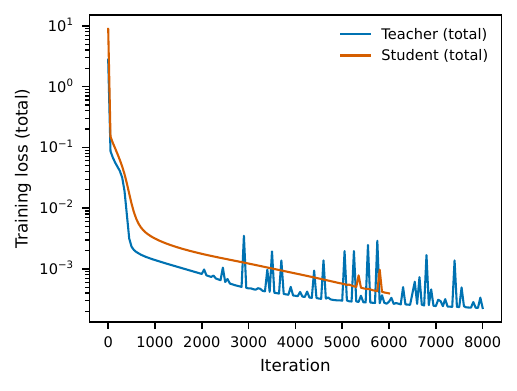}
    \caption{Total loss trajectories.}
    \label{fig:total_loss}
  \end{subfigure}

  \begin{subfigure}[b]{0.835\linewidth}
    \centering
    \raisebox{9pt}{
      \includegraphics[width=\linewidth]{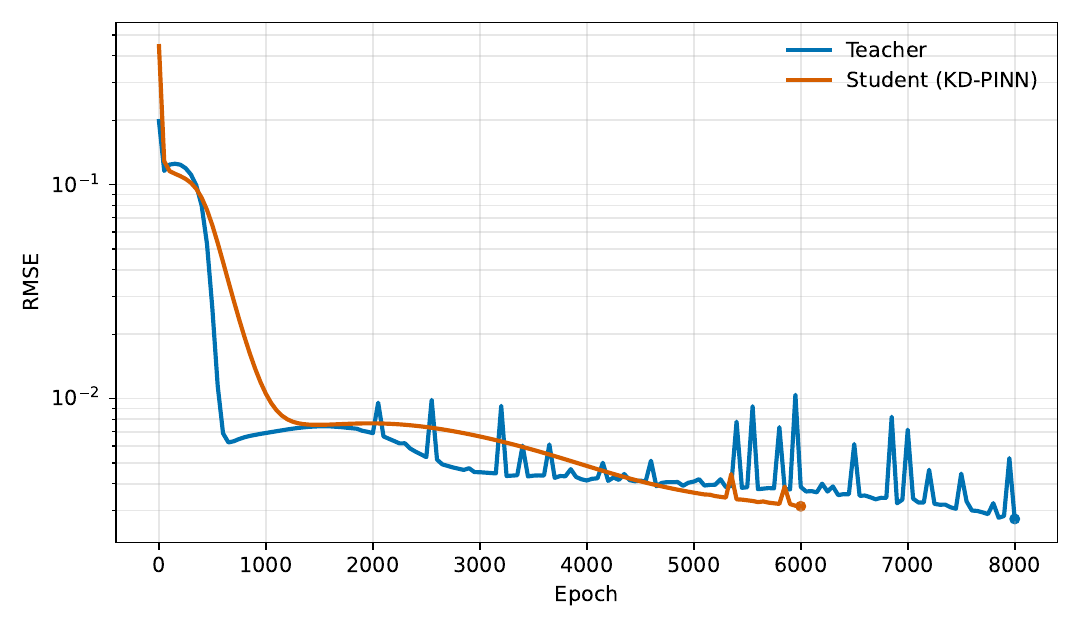}
    }
    \caption{RMSE vs.\ epochs.}
    \label{fig:rmse_epochs}
  \end{subfigure}

  \caption{
    Training dynamics of TeacherPINN and StudentPINN.
    (a) Total loss trajectories during training and (b) RMSE evolution over epochs.
  }
  \label{fig:training}
\end{figure*}

Consequently, in this early phase, the gradients, orient the global weight updates toward physics consistent teacher's space, inducing a better conditioning configuration. 
Thus, the PDE and boundary losses have smoother descent directions, resulting in accelerated 
convergence and a stabilization of physics-driven learning as shown by Figures \ref{fig:total_loss} and \ref{fig:rmse_epochs}.

After this alignment phase, once $u_{\theta}\!\approx\!u_{T}$ and $\mathcal{L}_{\mathrm{KD}}$ 
vanishes asymptotically, the optimization shifts toward physics-driven refinement 
($2000$–$4000$ iterations). $\mathcal{L}_{\mathrm{PDE}}$ and $\mathcal{L}_{\mathrm{BC}}$ 
progressively take over to enforce physical consistency.
Then a final physics-based convergence ($>4000$ iterations) ends the training.

This loss decomposition also provides an internal explanation of the RMSE dynamics observed 
in Figure~\ref{fig:rmse_epochs}, as the rapid decay of $\mathcal{L}_{\mathrm{KD}}$ and the 
balance induced in physics terms translates into the stabilized error trajectory of the 
StudentPINN.

\FloatBarrier
   
\paragraph{RMSE dynamics}
Figure~\ref{fig:rmse_epochs} compares the RMSE trajectories of the TeacherPINN and the distilled StudentPINN over training epochs. 
The error decreases rapidly from $10^{-1}$ to below $10^{-2}$ within the first 500 iterations. 
After this point, the teacher curve shows irregular oscillations of up to one order of magnitude ($10^{-3}$–$10^{-2}$), which indicates high gradient variance. 
In contrast, the student maintains a more uniform and monotonic decay, and keeps the RMSE below $10^{-2}$ during the entire training.  
This behavior confirms the regularizing effect of knowledge distillation: KD gradients stabilize the early learning phase, suppress stochastic fluctuations, and prevent the exploration of unstable minima.

\paragraph{Total loss trajectories}
Figure~\ref{fig:total_loss} compares the evolution of the total loss for the teacher and student networks, and confirms the regularizing role of the distillation. 

The teacher exhibits spikes after 2000 iterations, while the student converges uniformly, mainly due to conflicting gradients. In fact, during training, the gradients $\nabla_\theta \mathcal{L}_{\mathrm{PDE}}$ and $\nabla_\theta \mathcal{L}_{\mathrm{BC}}$ can temporarily compete.  Minimizing $\mathcal{L}_{\mathrm{PDE}}$ reduces the interior residuals but, since this constraint acts only on derivatives, it may globally shift $u_\theta$, thereby increasing the boundary error. 
Therefore, these gradients may point in divergent directions and the update
\begin{equation}
\Delta\theta = -\eta\,\nabla_\theta\mathcal{L}(\theta),
\end{equation}
can become large in norm, which produces the spikes observed in the teacher's total loss curve.

This effect becomes particularly apparent after the first phase (0-2000 iterations), where gradients were strong and both oriented toward the minimum, or at least the mean descent dominates their variance. As the error approaches the minimum, the network oscillates since weak conflicting gradients amplify the error due to the variance accross batches and numerical noise caused by sampling, floating-point precision and derivative computations. 

The Student’s total loss decreases more monotonically, since $\mathcal{L}_{\mathrm{KD}}$ first drives $u_\theta$ toward $u_T$ during the alignment phase, and thus, the PDE and boundary gradients become more coherent. In fact, even if they weaken near convergence, their correlation persists from the early phase of the training and constrains their convergence toward the minimum.

\noindent\textit{Remark.} Although the regularizing effect of KD is clearly observed experimentally, a rigorous mathematical characterization of how gradient correlation contributes to this regularization remains an open question.

\begin{figure*}[t]
  \centering
  \captionsetup[sub]{aboveskip=2pt,belowskip=1pt,justification=centering}

  \newcommand{\pairW}{0.87\textwidth}
  \newcommand{\pairH}{0.8\textwidth}
  \newcommand{\gap}{0.04\linewidth}
  \newcommand{\singleW}{0.87\textwidth}
  \newcommand{\singleH}{0.40\textwidth}

  \makebox[\textwidth][c]{%
    \begin{minipage}{\pairW}\centering
      \begin{subfigure}[t]{0.48\linewidth}
        \centering
        \includegraphics[height=\pairH,keepaspectratio,clip]{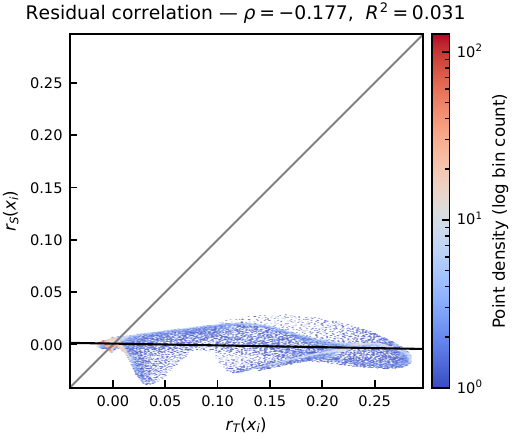}
        \caption{Residual correlation teacher–student.}
        \label{fig:f4_residual}
      \end{subfigure}\hspace{\gap}%
      \begin{subfigure}[t]{0.48\linewidth}
        \centering
        \includegraphics[height=\pairH,keepaspectratio,clip]{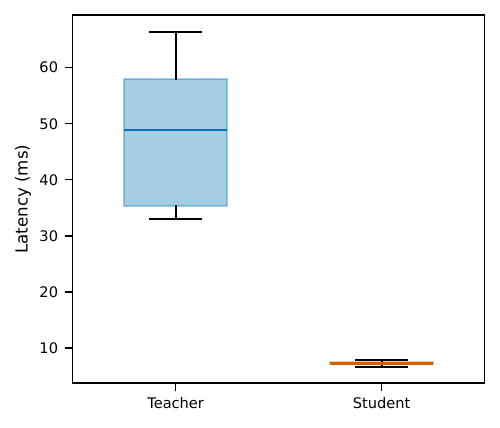}
        \caption{Inference latency comparison.}
        \label{fig:f5_latency}
      \end{subfigure}
    \end{minipage}%
  }

  \makebox[\textwidth][c]{%
    \begin{minipage}{\singleW}\centering
      \begin{subfigure}[t]{\linewidth}
        \centering
        \includegraphics[height=\singleH,keepaspectratio,clip]{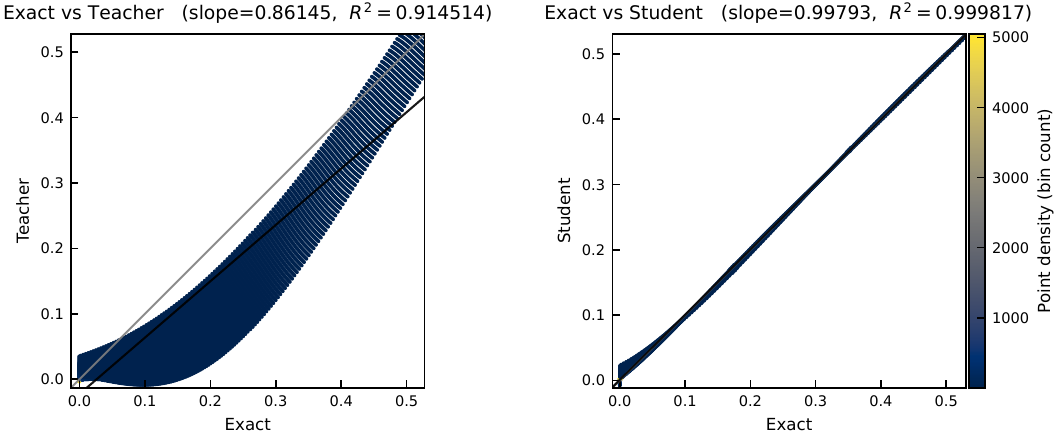}
        \caption{Calibration against the analytical Black--Scholes solution.}
        \label{fig:f6_calibration}
      \end{subfigure}
    \end{minipage}
  }

  \caption{Evaluation of StudentPINN performance on Black-Scholes.
  Top: (a) residual correlation and (b) inference latency
  Bottom: (c) calibration error.}
  \label{fig:BS-calibration}
  
\end{figure*}

\paragraph{Error correlation and calibration}
Figure~\ref{fig:f4_residual} shows that the residuals of the student are largely decorrelated from those of the teacher ($\rho=-0.177$), indicating that the student does not inherit the teacher's local errors. Moreover, the variance alignment is weak ($R^2=0.031)$: distillation suppresses oscillations and improves numerical stability. Figure~\ref{fig:f6_calibration} demonstrates the excellent calibration of the student with the analytical Black--Scholes solution ($R^2>0.999$), which demonstrates that distillation preserves the physical consistency imposed by the governing PDE.

\paragraph{Error profiles and spatial analysis}

\begin{figure*}[t]
  \centering
  \newlength{\colw}\setlength{\colw}{1\textwidth}
  \newlength{\pww}\setlength{\pww}{\colw/3}

  \begin{subfigure}[t]{\pww}
    \includegraphics[width=\linewidth]{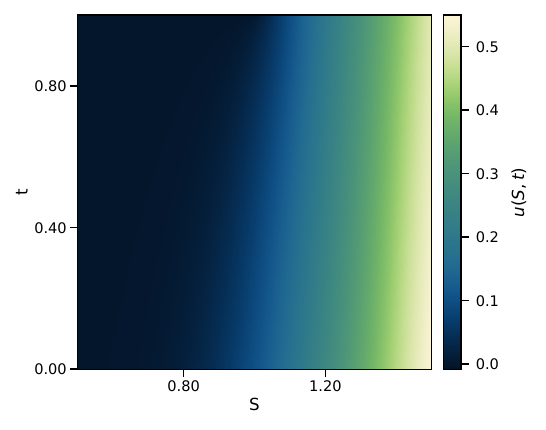}
    \caption*{(a) Exact}
    \label{fig:bsmapexact}
  \end{subfigure}%
  \begin{subfigure}[t]{\pww}
    \includegraphics[width=\linewidth]{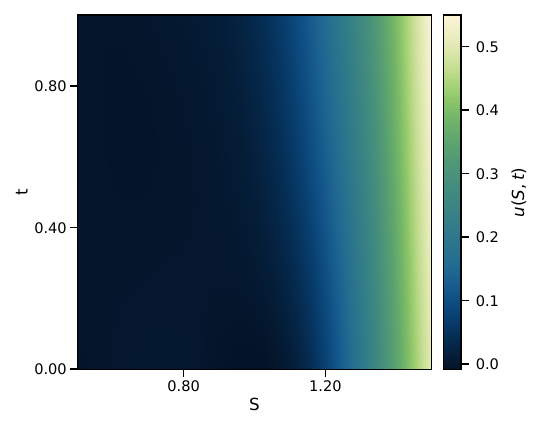}
    \caption*{(b) Teacher}
    \label{fig:bsmapteacher}
  \end{subfigure}%
  \begin{subfigure}[t]{\pww}
    \includegraphics[width=\linewidth]{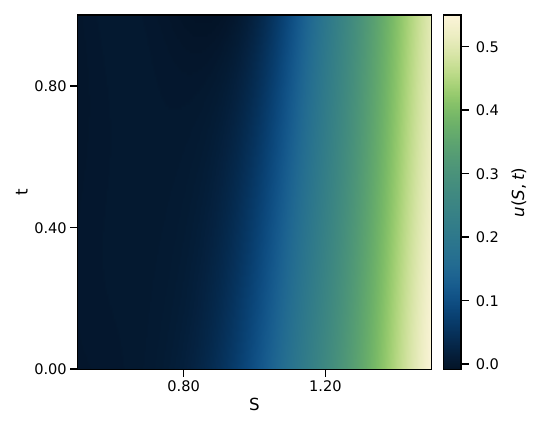}
    \caption*{(c) Student}
    \label{fig:bsmapstudent}
  \end{subfigure}

  \caption{
    Heatmaps of the Black–Scholes solution over the $(S,t)$ domain for the exact, teacher, and student models.
  }
  \label{fig:BS-evaluation}
\end{figure*}

Across the full $(S,t)$ domain (Fig.~\ref{fig:BS-evaluation}), the student reproduces the correct transition slope near the strike, while the teacher shows a slight inversion compared with the exact solution. This improvement suggests that distillation filters inconsistencies in the teacher’s local gradients.

\begin{figure*}[t]
  \centering
  \newlength{\pwg}\setlength{\pwg}{0.365\linewidth}

  \begin{subfigure}[t]{\pwg}
    \includegraphics[width=\linewidth]{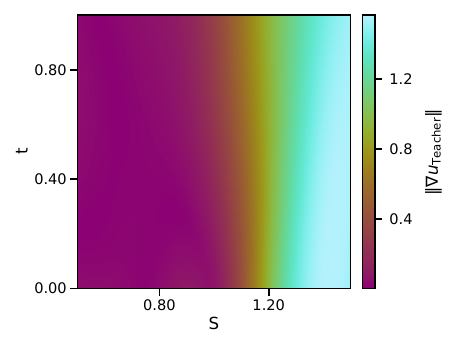}
    \caption*{(a) $\|\nabla u_{\mathrm{Teacher}}\|$}
  \end{subfigure}
  \begin{subfigure}[t]{\pwg}
    \includegraphics[width=\linewidth]{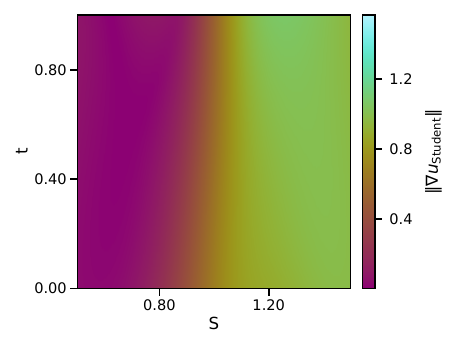}
    \caption*{(b) $\|\nabla u_{\mathrm{Student}}\|$}
  \end{subfigure}

  \caption{%
    Gradient-norm maps for the Teacher and Student PINNs 
    on the Black–Scholes equation.
  }
  \label{fig:BS-grad}
\end{figure*}

Figure~\ref{fig:BS-grad} shows a more homogeneous $\|\nabla u(S,t)\|_2$ field and smaller errors for the distilled model. It confirms that distillation regularizes the updates and stabilizes the optimization process.
The gradient variations of the teacher increase more than those of the student above the strike ($S=1$), indicating that the teacher’s errors in that region propagate toward the boundary. This demonstrates that knowledge distillation effectively limits the spread of local errors in a diffusion model.

\paragraph{Role of knowledge distillation in leading to reduced inference latency}

The ablation using identical architectures (\texttt{[2,50,50,50,1]}) clearly isolates the source of latency reduction. 
With the same layers, depth, and activations the distilled model achieves a $45\%$ lower RMSE than its non-distilled counterpart, while latency remains unchanged ($\Delta t_{\mathrm{inf}}<3\%$).  
This confirms that the latency gain in compressed models is attributable to the reduced depth, whereas the maintained accuracy results from the regularizing effect of distillation.

\FloatBarrier
\subsection{Tuning for extrapolation and robustness: KD-PINN$^{+}$}
This section investigates the extrapolation performance of KD--PINN and demonstrates that it can be enhanced through targeted parameter tuning based on an examination of the training dynamics.

\subsubsection{Evaluation protocol}
We assess boundary-proximal generalization on five regions outside or at the edge of the train box $[0.5,1.5]\times[0,1]$:
\emph{mild right} ($S\!\in[1.6,2.0]$), \emph{moderate right} ($S\!\in[2,3]$), \emph{hard right} ($S\!\in[3,5]$), \emph{left} ($S\!\in[0.2,0.49]$), and \emph{diag} ($S\!\in[1.6,3],\,t\!\in[0.6,1]$).
On each box, predictions are evaluated on a fixed $200\times80$ grid (same input scaling as in-domain) and compared pointwise to the analytical Black--Scholes solution.
We report RMSE and relative $L^2$ error, and examine the pointwise absolute error 
$|{\rm error}(S,t)| = |u_{\text{pred}}(S,t) - u_{\text{ref}}(S,t)|$ 
as a function of the $L_\infty$ distance to the train box 
$d_\infty((S,t),D)=\max\!\big(\max(0,\,0.5\!-\!S,\,S\!-\!1.5),\,\max(0,\,0\!-\!t,\,t\!-\!1)\big)$, $(S,t)\!\in\![0,5]\!\times\![0,1]$.

\subsubsection{Extrapolation behavior and experimental tuning}
\begin{figure*}[htbp]
    \begin{center}
  \scalebox{1}{
  \begin{minipage}{1\textwidth}
    \begin{subfigure}[t]{0.5\linewidth}
      \includegraphics[width=\linewidth]{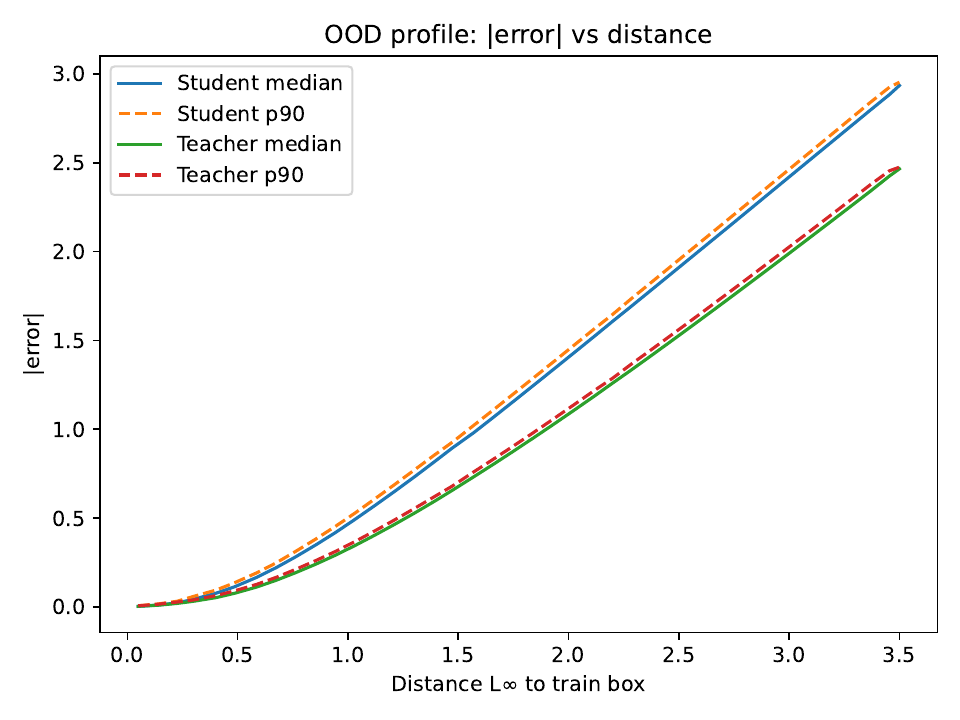}
      \caption{$|{\rm error}|$ vs distance ($L_\infty$) — baseline KD--PINN.}
      \label{fig:ood-linfini-base}
    \end{subfigure}
    \begin{subfigure}[t]{0.5\linewidth}
      \includegraphics[width=\linewidth]{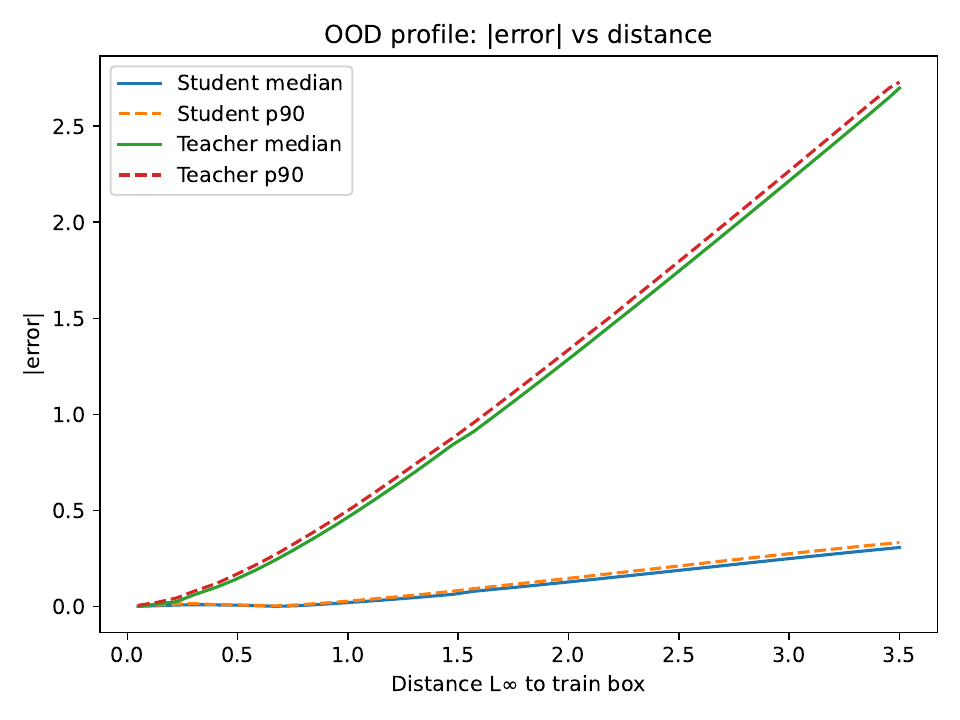}
      \caption{$|{\rm error}|$ vs distance ($L_\infty$) — KD--PINN$^{+}$.}
      \label{fig:ood-linfini-mitig}
    \end{subfigure}
  \end{minipage}
  }
  \end{center}
  \caption{Error vs distance ($L_\infty$) for baseline and KD–PINN$^{+}$.}
  \label{fig:ood-distance-only}
\end{figure*}

The baseline KD--PINN displays a continuous degradation pattern when inputs move away from the calibration range. 
Figure~\ref{fig:ood-linfini-base} shows a continuous increase in median absolute error with distance $L_\infty$, that indicates a progressive degradation in generalization outside the training domain. After a distance $d_\infty \!\approx\!3$, the student's error exceeds that of the teacher by approximately $20$–$25\%$, and thus reflects a stronger dependence on training domain features. This loss of generalization results from the three main causes described below.

\paragraph{Sampling bias and boundary sparsity}
First, the Sobol sampling density $p_{\text{train}}(S,t)$ concentrates collocation points inside $\Omega_{\rm train}$, leaving boundary regions underrepresented. 
The physical loss is estimated over a discrete set of collocation points 
$\{x_i\}_{i=1}^N$ sampled from $p_{\text{train}}$, 
so that 
\(
\widehat{\mathcal{L}}_{\text{phys}}=\tfrac{1}{N}\sum_i |\mathcal{N}[u_\theta(x_i)]|^2.
\)
Under the i.i.d. assumption, this empirical loss converges in probability to its continuous counterpart,
\begin{equation}
\mathcal{L}_{\text{phys}}
=\int_{\Omega_{\text{train}}}\!|\mathcal{N}[u_\theta(S,t)]|^2\,p_{\text{train}}(S,t)\,dS\,dt,
\end{equation}
as $N\!\to\!\infty$, where the Sobol-based density is defined with respect to the interior volume measure $dS\,dt$. 
Because this measure distributes samples uniformly inside the domain, the relative fraction of points within a thin boundary layer decreases proportionally to its thickness. 

Informed sampling \citep{Cuomo2022SciAdv} mitigates this bias by introducing a weighted density 
$p_{\text{train}}^{\mathrm{inf}}(S,t)=w(S,t)/\!\int_{\Omega_{\text{train}}}\!w$,
where $w(S,t)$ increases near $\partial\Omega_{\text{train}}$ or in high-residual zones.
The modified loss
\begin{equation}
\mathcal{L}_{\text{phys}}^{\mathrm{inf}}(\theta)
=\frac{\int_{\Omega_{\text{train}}}|\mathcal{N}[u_\theta(S,t)]|^2\,w(S,t)\,dS\,dt}
{\int_{\Omega_{\text{train}}}w(S,t)\,dS\,dt}.
\label{eq:L_phys_inf}
\end{equation}
balances the contribution of interior and boundary regions, improving gradient conditioning and extrapolation stability.
In the present experiments, the weighting function is derived from the teacher’s residual field,
\begin{equation}
w(S,t)=1+\eta\,\frac{|\mathcal{N}[u_T(S,t)]|}{\max_{\Omega_{\text{train}}}|\mathcal{N}[u_T(S,t)]|},
\label{eq:residual-weight}
\end{equation}
with $\eta\!\in\![0.5,1.0]$ controlling the degree of bias toward high-error regions. Where the teacher solution is accurate, $|\mathcal{N}[u_T(S,t)]|$ is small and $w(S,t)\!\approx\!1$,
in contrast, high residuals near boundaries or steep gradients increase $w(S,t)$ and draw more training focus to these difficult areas.
This sampling is well suited to the KD setting where the teacher already identifies the most challenging regions.

\paragraph{Loss function sensitivity}
Second, the quadratic mean-squared loss $\mathcal{L}_{\text{MSE}}$ amplifies the impact of rare but large residuals, which produces unstable gradient updates when $|u_{\text{pred}}-u_{\text{ref}}|\!\gg\!1$. 
The Huber loss~\citep{Huber1964} transitions from quadratic to linear growth beyond a threshold $\delta$, reducing the dominance of large residuals:
\begin{equation}
\mathcal{L}_\delta(r) =
\begin{cases}
\tfrac{1}{2}r^2, & |r|\le\delta,\\[4pt]
\delta(|r|-\tfrac{1}{2}\delta), & \text{otherwise.}
\end{cases}
\label{eq:Huber}
\end{equation}

\paragraph{High curvature of the loss landscape}
Finally, compact networks with steep learning-rate schedules exhibit high curvature in the loss landscape,
corresponding to sharp minima~\citep{Keskar2017LargeBatch} and insufficient robustness to data shifts.
In the out of domain (OOD) regime, the input distribution $p_{\text{test}}(x)$ differs from
$p_{\text{train}}(x)$, which induces a shift in the optimal parameters.
The new minimizer $\theta^\star_{\text{OOD}}$ can be viewed as a perturbation
of the in-domain optimum $\theta^\star$, i.e.
$\theta^\star_{\text{OOD}} = \theta^\star + \Delta$,
where $\Delta$ denotes the parameter displacement induced by the distribution shift.
Substituting this perturbation into the Taylor expansion yields
\(
\nabla_\theta \mathcal{L}(\theta+\Delta)
\simeq \nabla_\theta \mathcal{L}(\theta) + H\Delta
= \nabla_\theta \mathcal{L}(\theta)
+ \sum_i \lambda_i (v_i^\top \Delta)\,v_i,
\)
where $H=\nabla_\theta^2\mathcal{L}$ is the Hessian of the loss and $(\lambda_i,v_i)$ its eigenpairs, with $H v_i = \lambda_i v_i$ and $\lambda_i>0$ near a local minimum.
When the Hessian $H$ has large eigenvalues (high curvature),
even a small parameter shift $\Delta$
produces a large gradient change and thus a sharp increase in loss.

Curriculum learning~\citep{Bengio2009Curriculum} addresses this issue by progressively increasing task complexity
through a scheduling variable $c(t)$ that regulates the contribution of different loss components during training.  This progressive weight moderates the gradient flow and avoids abrupt updates, which reduces the local curvature of the loss landscape. Thus, the model becomes less sensitive to parameter change.
In our experiments, we adopted a simple linear schedule
\(
c(t)=\min\!\left(1,\,\frac{t}{T_c}\right),
\)
where $T_c$ denotes the characteristic number of training iterations required to reach the full weighting of the physical constraint.
In the present experiments, we set $T_c = 0.2\,T_{\max}$, corresponding to a linear increase of the physics weighting during the first 1000 iterations.
During this phase, the total loss is expressed as
\(
\mathcal{L}_{\text{total}}(\theta,t)
=(1-c(t))\,\mathcal{L}_{\text{data}}(\theta)
+c(t)\,\mathcal{L}_{\text{phys}}(\theta),
\)
so that the physical loss remains underweighted while the network learns to reproduce the general solution structure, through knowledge distillation.

\subsubsection{Performance assessment}

\begin{figure*}[t]
  \centering

  \newcommand{\oodscale}{1}

  \begin{adjustbox}{center,scale=\oodscale}
  \begin{minipage}{1\textwidth} 

  \begin{center}
  \begin{subfigure}[t]{0.48\linewidth}
    \includegraphics[width=\linewidth]{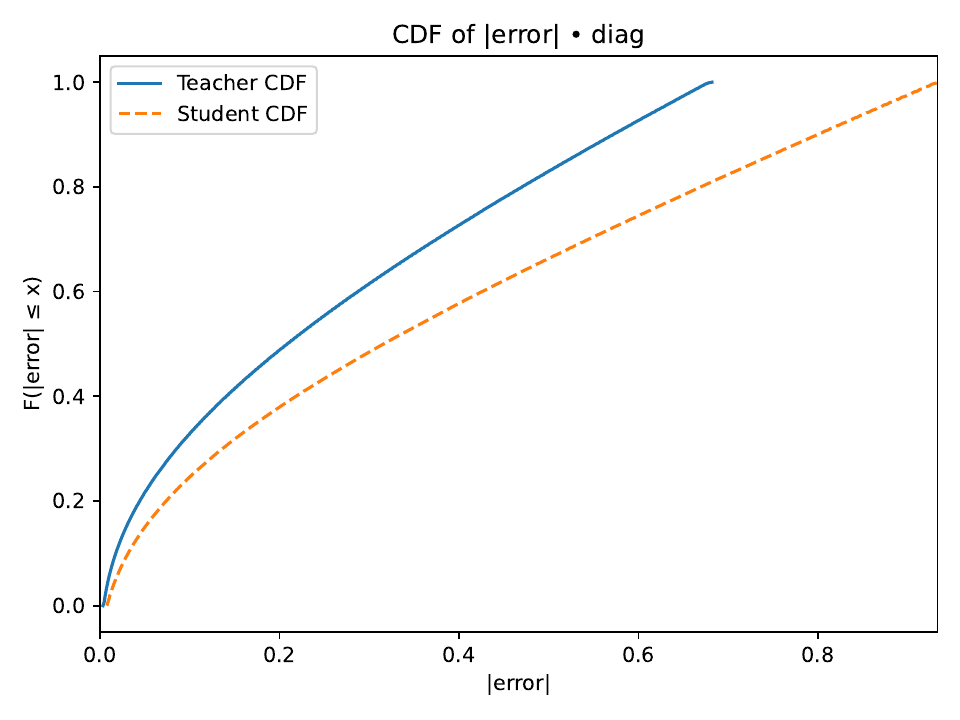}
    \caption{CDF of absolute errors (\textit{diag}) — baseline.}
    \label{fig:ood-cdf-diag-base}
  \end{subfigure}
  \begin{subfigure}[t]{0.48\linewidth}
    \includegraphics[width=\linewidth]{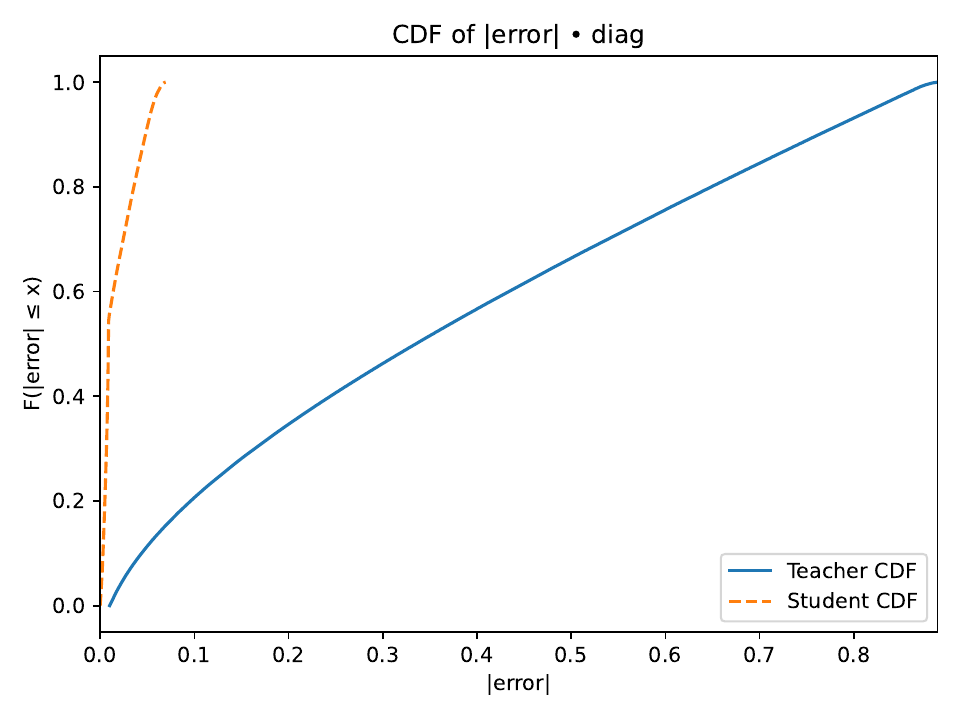}
    \caption{CDF of absolute errors (\textit{diag}) — KD--PINN$^{+}$.}
    \label{fig:ood-cdf-diag-mitig}
  \end{subfigure}
  \end{center}

  \begin{center}
  \begin{subfigure}[t]{0.4\linewidth}
    \includegraphics[width=\linewidth]{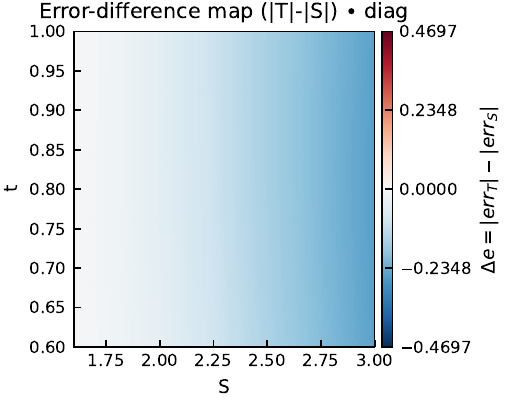}
    \caption{Error difference map (\textit{diag}) — baseline.}
    \label{fig:ood-err-diag-base}
  \end{subfigure}
  \begin{subfigure}[t]{0.4\linewidth}
    \includegraphics[width=\linewidth]{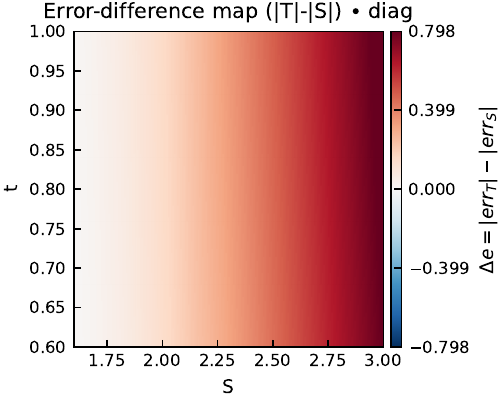}
    \caption{Error difference map (\textit{diag}) — KD--PINN$^{+}$.}
    \label{fig:ood-err-diag-mitig}
  \end{subfigure}
  \end{center}

  \caption{
  Generalization analysis for the baseline and the improved KD--PINN$^{+}$.
  (a–b) Cumulative distributions of absolute errors 
  (c–d) Error difference maps}
  \label{fig:ood-panel}
  \end{minipage}
  \end{adjustbox}
\end{figure*}

The stabilized KD--PINN$^{+}$ achieves markedly smoother error growth, as shown in Figure~\ref{fig:ood-linfini-mitig}, with reduced sensitivity to domain shift. 
Figures~\ref{fig:ood-cdf-diag-mitig} and ~\ref{fig:ood-err-diag-mitig} show that in the diagonal region, the most difficult OOD case, the Student$^{+}$ attains lower errors than the teacher, confirming its improved extrapolation performance.

\begin{table*}[hbtp]
\centering
\caption{Extrapolation performance comparison between KD-PINN$^{+}$ (Student) and Teacher.
Ratios below~1 indicate better performance of KD-PINN$^{+}$.}
\label{tab:ood-kdpinnplus}
\small
\begin{tabular}{lrrrrrr}
\toprule
\multirow{2}{*}{Region} & \multicolumn{3}{c}{RMSE} & \multicolumn{3}{c}{rel-$L^2$} \\
\cmidrule(lr){2-4}\cmidrule(lr){5-7}
 & Teacher & Student$^{+}$ & \textbf{S$^{+}$/T} & Teacher & Student$^{+}$ & \textbf{S$^{+}$/T} \\
\midrule
left            & 1.230e$-$02 & 8.154e$-$03 & \textbf{0.663} & 1.471e$+$04 & 9.754e$+$03 & \textbf{0.663} \\
mild\_right     & 7.678e$-$02 & 8.447e$-$03 & \textbf{0.110} & 9.219e$-$02 & 1.014e$-$02 & \textbf{0.110} \\
moderate\_right & 5.137e$-$01 & 3.577e$-$02 & \textbf{0.070} & 3.310e$-$01 & 2.305e$-$02 & \textbf{0.070} \\
hard\_right     & 1.834e$+$00 & 2.018e$-$01 & \textbf{0.110} & 5.954e$-$01 & 6.552e$-$02 & \textbf{0.110} \\
diag            & 4.541e$-$01 & 2.542e$-$02 & \textbf{0.056} & 3.311e$-$01 & 1.854e$-$02 & \textbf{0.056} \\
\midrule
Aggregate & --          & --          & \textbf{0.202} & --          & --          & \textbf{0.202} \\
\bottomrule
\end{tabular}
\end{table*}

Table~\ref{tab:ood-kdpinnplus} shows that across all regimes, Student$^{+}$ performs better than the teacher out of domain. OOD errors are reduced by 89–94\%, with Student/Teacher ratios $\leq0.11$ in the hardest boxes and $\approx0.20$ overall.
These results demonstrate that the proposed mitigation pipeline effectively enhances robustness and generalization. The extrapolation capacity of knowledge-distilled PINNs is then flexible and can be tuned through suitable design choices.

\subsection{Generalization to other PDEs}
\label{sec:cross-pde}

To assess the consistency across PDEs the student is trained on three canonical benchmark problems of increasing complexity: the one-dimensional Burgers and Allen--Cahn equations, and the two-dimensional Navier--Stokes equations.

\subsubsection{PDE setup}
The viscous Burgers equation is considered to study sensitivity to nonlinear transport and high-gradient regimes. 
It models nonlinear advection–diffusion dynamics with steep gradients and shock formation, and is defined as
\begin{subequations}
\label{eq:burgers_full}
\begin{align}
u_t + u\,u_x - \nu\,u_{xx} &= 0,
\qquad (x,t)\in[0,1]^2, \label{eq:burgers_eq}\\[3pt]
u(x,0) &= -\sin(\pi x), \label{eq:burgers_ic}\\[3pt]
u(0,t) &= 0, \qquad u(1,t) = 0. \label{eq:burgers_bc}
\end{align}
\end{subequations}

The Allen--Cahn equation describes reaction–diffusion and phase-transition phenomena under strong stiffness, and provides a test of stability for small-scale features. 
It is defined as
\begin{subequations}
\label{eq:allen_full}
\begin{align}
u_t - \nu\,u_{xx} + u^3 - u &= 0,
\qquad (x,t)\in[-1,1]\times[0,1], \label{eq:allen_eq}\\[3pt]
u(x,0) &= x^2\cos(\pi x), \label{eq:allen_ic}\\[3pt]
u(-1,t) &= 0, \qquad u(1,t) = 0. \label{eq:allen_bc}
\end{align}
\end{subequations}

Finally, we consider the incompressible Navier--Stokes equations on the periodic domain 
$\Omega = (0, 2\pi)^2$ over $t \in (0, 1]$, with viscosity $\nu = 10^{-2}$. 
They are defined as
\begin{subequations}
\label{eq:ns_full}
\begin{align}
\partial_t \mathbf{u} + (\mathbf{u}\!\cdot\!\nabla)\mathbf{u} + \nabla p - \nu \Delta \mathbf{u} &= \mathbf{0}, 
\qquad (\mathbf{x}, t) \in \Omega \times (0, 1], \label{eq:ns_momentum}\\[3pt]
\nabla\!\cdot\!\mathbf{u} &= 0, \label{eq:ns_incompressible}
\end{align}
\end{subequations}
where the velocity field $\mathbf{u} = (u, v)$ and the pressure field $p$ are $2\pi$-periodic in both $x$ and $y$. 
This configuration admits the exact Taylor--Green vortex solution,
\begin{subequations}
\label{eq:taylor_green_full}
\begin{align}
u(x, y, t) &= \cos x\,\sin y\,e^{-2\nu t}, \label{eq:taylor_green_u}\\[3pt]
v(x, y, t) &= -\sin x\,\cos y\,e^{-2\nu t}, \label{eq:taylor_green_v}\\[3pt]
p(x, y, t) &= \tfrac{1}{4}\!\left(\cos 2x + \cos 2y\right)e^{-4\nu t}. \label{eq:taylor_green_p}
\end{align}
\end{subequations}
Periodic boundary conditions are applied on $\partial\Omega$, 
and the initialization is taken from the exact field at $t = 0$,
\begin{subequations}
\label{eq:taylor_ic_full}
\begin{align}
\mathbf{u}(x, y, 0) &= (\cos x\,\sin y,\; -\sin x\,\cos y), \label{eq:taylor_ic_u}\\[3pt]
p(x, y, 0) &= \tfrac{1}{4}\!\left(\cos 2x + \cos 2y\right). \label{eq:taylor_ic_p}
\end{align}
\end{subequations}
This problem represents a coupled, multi-field flow in which velocity and pressure are interdependent through the incompressibility constraint, the network must therefore learn a constrained vector structure, where errors in one component can propagate through the coupling term $\nabla p$.
Navier-Stokes thus allows the assessment of the framework's ability to maintain physical coherence across variables.

\subsubsection{Experimental setup}
Hyperparameters and sampling follow the Black--Scholes setup. 
Architectures: Teacher $(64\times4)$ and compact Student $(20\times3)$, both using $\tanh$ activations. 
For Navier--Stokes, slight adaptations handle the coupled fields $(u,v,p)$ and higher dimensionality: 
the Student uses SiLU activations for smoother gradients, a width of~32 to capture cross-channel effects, and a higher distillation temperature ($\tau{=}2$) for stability under periodic boundaries. 
Models were compiled with \texttt{torch.compile} and \texttt{TorchScript} to reduce kernel overhead and inference latency. 
Latency was measured on CPU over $2\times10^4$ batched inputs, reported as the median of 100 runs.

\begin{figure*}[htbp]
  \centering
  \scalebox{0.9}{
  \begin{minipage}{1.0\linewidth}
    \centering

    \begin{subfigure}[t]{0.48\linewidth}
      \centering
      \includegraphics[width=\linewidth]{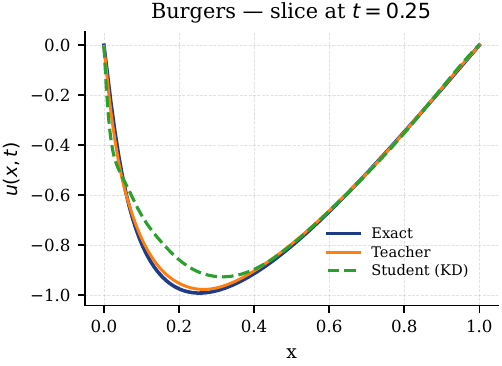}
      \caption{Burgers --- slice at $t=0.25$.}
      \label{fig:slice-burgers}
    \end{subfigure}\hfill
    \begin{subfigure}[t]{0.48\linewidth}
      \centering
      \includegraphics[width=\linewidth]{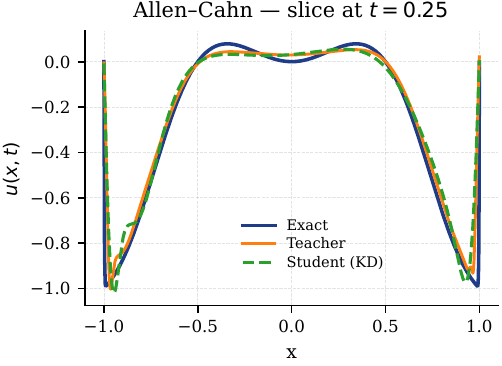}
      \caption{Allen--Cahn --- slice at $t=0.25$.}
      \label{fig:slice-allencahn}
    \end{subfigure}

    \begin{subfigure}[hbtp]{0.98\linewidth}
      \centering
      \includegraphics[width=\linewidth]{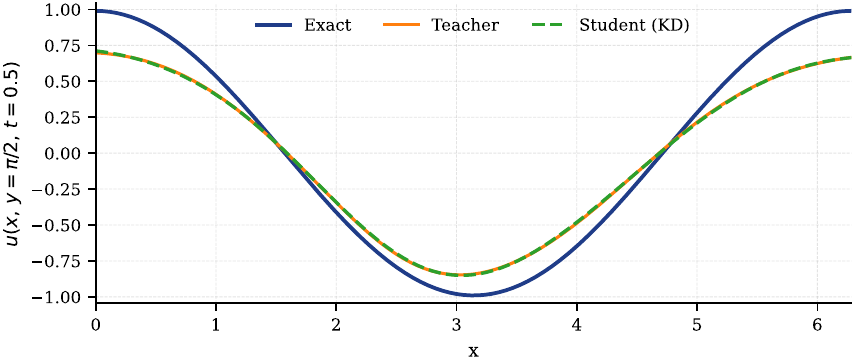}
      \caption{Navier--Stokes (latency-opt) --- slice at $y=\pi/2$, $t=0.5$.}
      \label{fig:slice-ns}
    \end{subfigure}

    \caption{
    One-dimensional solution profiles comparing exact, teacher, and student solutions
    across nonlinear PDEs.
    }
    \label{fig:slices-multiPDE}
  \end{minipage}
  } 

  \vspace{0.6\baselineskip} 

  \begin{minipage}{1.0\linewidth}
    \centering
    \small
    \captionof{table}{Cross-PDE evaluation of KD--PINN. Latencies measured on CPU inference.}
    \label{tab:crosspde}
    \setlength{\tabcolsep}{5pt}
    \begin{tabular}{llccccc}
    \toprule
    PDE & Config. & RMSE$_{\mathrm{T}}$ & RMSE$_{\mathrm{S}}$ &
    Lat.$_{\mathrm{T}}$ [ms] & Lat.$_{\mathrm{S}}$ [ms] & Speed-up \\
    \midrule
    Allen--Cahn & Baseline & \num[round-mode=figures, round-precision=3]{9.13e-2} & \num[round-mode=figures, round-precision=3]{1.00e-1} & 25.7 & 5.3 & $\times$4.87 \\
    Burgers & Baseline & \num[round-mode=figures, round-precision=3]{3.49e-2} & \num[round-mode=figures, round-precision=3]{4.16e-2} & 32.1 & 4.6 & $\times$6.92 \\
    Navier--Stokes & Baseline & \num[round-mode=figures, round-precision=3]{1.31e-1} & \num[round-mode=figures, round-precision=3]{1.66e-1} & 26.8 & 6.9 & $\times$3.90 \\
    Navier--Stokes & Latency-opt. & \num[round-mode=figures, round-precision=3]{1.31e-1} & \num[round-mode=figures, round-precision=3]{1.30e-1} & 19.4 & 4.1 & $\times$4.76 \\
    \bottomrule
    \end{tabular}
  \end{minipage}

\end{figure*}

\paragraph{Tuning for the 2D case of Navier-Stokes}
For Navier--Stokes, the baseline Student is slightly less accurate (RMSE $1.66\!\times\!10^{-1}$ vs.\ $1.31\!\times\!10^{-1}$) with a $3.9\times$ speed-up, but a latency-optimized variant (SiLU, width~32, three layers, $\tau{=}2$, TorchScript/compile) restores parity with the teacher and raises speed-up to $4.8\times$ (Table~\ref{tab:crosspde}). 
The decrease of accuracy results from limited spectral bandwidth: compact $\tanh$ networks act as low-pass filters with Fourier amplitudes decaying as $|\widehat{f_\theta}(\omega)|\!\sim\!e^{-c|\omega|}$~\citep{Rahaman2019FPrinciple}, which restricts the representation of high-frequency modes needed for vortical cores and pressure–velocity coupling. Indeed, since the vorticity field is defined as 
\(
\omega_v = \nabla \times u
\)
and
\(
\nabla u = \int_{-\infty}^{\infty} (i\omega)\,\hat{u}(\omega,t)\,e^{i\omega x}\,d\omega,
\)
high vorticity corresponds to large \(|\omega|\) components, linking vortical structures to the high-frequency content of \(u(x,t)\).
Any neural network can be written as 
\(
u_\theta(x)=\int_{\mathbb{R}^d}\widehat{u_\theta}(\omega)e^{i\omega\cdot x}d\omega,
\)
so missing high-$\omega$ components imply $|\widehat{u_\theta}(\omega)|\!\approx\!0$, leaving no parameter directions to capture fine scales.  
During training, the optimizer compensates by increasing weight magnitudes, pushing activations into the saturation regime of $\tanh(z)$ ($\tanh(z)\to\pm1$ for $|z|\gg1$), where $\tanh'(z)=1-\tanh^2(z)\!\to\!0$. 
Since $\tfrac{\partial u_\theta}{\partial W}=\tanh'(z)\,x$, small perturbations of $W$ can shift $z=Wx+b$ across steep transition regions, causing disproportionate and irregular variations in $u_\theta(x)$.
Indeed, as the Jacobian $J = \partial u_\theta / \partial \theta$ grows in norm, the Hessian $H \!\approx\! J^{\top}J$ grows in magnitude which results in a steepened loss landscape. As $\nabla_\theta \mathcal{L} = J^{\top}(u_\theta - u^{*})$, this amplification of curvature directly induces the increased sensitivity of the loss to parameter perturbations, and justifies the expansion of the network architecture. 
Moreover, when $\tanh$ saturates, the CPU keeps spending computations on nearly constant activations, which lowers computational efficiency and increases latency, as observed in CPU inference benchmarks~\citep{Pochelu2022DLInferenceBenchmark}. In contrast, the SiLU activation $\mathrm{SiLU}(z)=z\,\sigma(z)$ is non-saturating for $z>0$~\citep{Elfwing2018SiLU} and preserves informative gradients over a wider dynamic range. Thus, increasing the representation capacity and adapting the activation function (SiLU, width 32, three layers) jointly improves the accuracy and latency of the student.

\subsubsection{Results}
Table~\ref{tab:crosspde} summarizes accuracy and latency. 
For Burgers and Allen-Cahn, the student shows a decrease in accuracy on the order of $10\%$ compared to the teacher, in exchange for a significant increase in inference speed.
In contrast, the Navier--Stokes latency-optimized configuration shows that,
with adequate tuning, the student achieves an accuracy 
that is approximately $1\%$ better in terms of root mean square error (RMSE) than the teacher.
We note that the reported error levels do not reflect intrinsic limitations of the proposed framework. Complementary experiments indicate that further improvements in accuracy are possible in more favorable training configurations, while preserving the same qualitative acceleration behavior.

In the one-dimensional slices of the Figure~\ref{fig:slices-multiPDE}, a quantitative agreement with the analytical solutions can be observed, noting that the student retains both the amplitude and the phase for the different regimes.
In particular, for the Navier--Stokes case, in Fourier space, the physical invariants (energy, enstrophy, and vorticity) depend solely on the spectral amplitude and phase distribution of $\hat{u}(\omega,t)$. Their preservation directly implies the conservation of the underlying flow physics.
This confirms that the model retains the underlying physics even under coupling, higher dimensionality, and nonlinear effects.

\section{Analysis of the acceleration ratio}
\label{sec:analysis-ratio}
In order to complement these empirical observations, the theoretical ceiling of acceleration achievable through knowledge distillation can be formalized.

Let \(S = \tfrac{\text{latency}_{\text{Teacher}}}{\text{latency}_{\text{Student}}}\) denote the speed-up factor. 
Three complementary factors together bound $S$.
At the algorithmic level, the inference cost of a fully connected network scales with its floating-point operations. 
If $R_{\text{FLOPs}}$ is the ratio of teacher to student FLOPs and $d_l$ denotes the number of neurons of layer $l$ of a MLP of L layers, 
\(
S \le R_{\text{FLOPs}} = \frac{\sum_{l=1}^{L} d_{l-1}d_l}{\sum_{l=1}^{L'} d'_{l-1}d'_l},
\) since no compression can exceed the reduction in arithmetic complexity here expressed through the total number of multiply–accumulate operations, i.e.
the canonical FLOP-count expression for fully connected layers~\citep{Han2016DeepCompression, Blalock2020Pruning}.
From Amdahl’s law~\citep{Amdahl1967, Hill2008Amdahl}, the existence of non-accelerable runtime costs $f$ (kernel launches, CPU/GPU memory transfers) further bounds the gain:
\(
S \le \frac{1}{f + (1-f)/R_{\text{FLOPs}}},
\)
which implies a saturation at $S \to 1/f$ even for large $R_{\text{FLOPs}}$. This component highlights that architectural compression alone is insufficient once the runtime is dominated by memory and scheduling costs.
\noindent
Finally, the Roofline model~\citep{Williams2009Roofline} defines a hardware-imposed limit governed by arithmetic intensity and memory bandwidth. 
The effective performance of a model is bounded by 
$\text{Perf} = \min(P_{\text{peak}},\, \text{AI}\!\times\!\text{BW})$, 
where $P_{\text{peak}}$ denotes the peak arithmetic throughput of the processor, 
$\text{AI}$ the arithmetic intensity (FLOPs per byte transferred), 
and $\text{BW}$ the effective memory bandwidth between compute units and memory. 
$S$ can be expressed as 
\(
S = \frac{t_T}{t_S} 
  = \frac{F_T / P_T}{F_S / P_S}
  = R_{\text{FLOPs}} \times \frac{P_S}{P_T},
\)
where $F$ is the total number of floating-point operations and $P$ the achieved performance (FLOPs/s), since $P_T = \min(P_{\text{peak}},\, \text{AI}_T \times \text{BW})$ and 
$P_S = \min(P_{\text{peak}},\, \text{AI}_S \times \text{BW})$,

\begin{equation}
S \le R_{\text{FLOPs}}
   \times \frac{\min(P_{\text{peak}},\, \text{AI}_S\text{BW})}
                {\min(P_{\text{peak}},\, \text{AI}_T\text{BW})}
   \;\Rightarrow\;
   S \le R_{\text{FLOPs}}\!\times\!\min\!\Big(1,\,\tfrac{\text{AI}_S}{\text{AI}_T}\Big).
\end{equation}
This hardware limit indicates that acceleration saturates when the student reaches the memory-bandwidth ceiling 
($\text{AI}_S\,\text{BW} < P_{\text{peak}}$), 
that is, when its arithmetic intensity becomes lower than that of the teacher and the computation becomes memory-bound.

Combining these yields the global bound
\begin{equation}
S_{\max} \;\le\; \min\!\left(
R_{\text{FLOPs}},\,
\frac{1}{\,f + (1-f)/R_{\text{FLOPs}}\,},\,
R_{\text{FLOPs}}\!\cdot\!\min\!\Big(1,\,\tfrac{\text{AI}_S}{\text{AI}_T}\Big)
\right).
\end{equation}

For our setup, the FLOP ratio between the teacher and student 
(\([2,50,50,50,1]\) vs.\ \([2,20,20,20,1]\)) is about $R_{\text{FLOPs}}\!\approx\!6$. 
Accounting for activations and ancillary operations, the effective ratio is conservatively bounded by~10. 
On standard T4 GPU and Xeon CPU hardware, where kernel-launch and memory-transfer overheads represent $f\!\in[0.02,0.05]$ of total latency, 
the resulting theoretical bound is $S_{\max}\!\in[7,12]$. 
This agrees with the maximal measured $6.9\times$ acceleration, and confirms that the observed gain already nears the hardware–algorithmic limit for this type compact MLPs.

Beyond this range, compact MLPs used in PINNs usually have low intensity 
($AI\!\approx\!0.1$), resulting in effective utilization ($P_{achieved}/P_{Roofline}$) below $1\text{–}2\%$ of $P_{\text{peak}}$, as confirmed by recent CPU/GPU benchmarks ~\citep{Pochelu2022DLInferenceBenchmark, GomezLuna2023LowAI}.
In an idealized compute-bound regime, the Roofline model yields the theoretical upper bound
\begin{equation}
S_{\max} = \frac{P_{\text{peak}}}{P_{\text{current}}} \approx \frac{1}{\text{Utilization}} \in [50,100],   
\end{equation}
meaning that, in an ideal hardware regime with full resource utilization, up to two orders of magnitude in additional speed-up could in theory be reached. In practice, this theoretical potential remains largely unattained because compact MLPs used in PINNs exhibit low arithmetic intensity and fragmented execution (each layer being launched as a separate kernel), spending more time moving data than performing computations during inference.
They thus operate in a memory-bound regime, with most compute units under-utilized despite significant power draw. Approaching this limit would require increasing arithmetic intensity through wider or fused layers, improving kernel fusion and vectorization, replacing transcendental activations such as $\tanh$ with lightweight smooth functions (SiLU, GELU), optimizing memory traffic via mixed-precision and prefetching, and leveraging specialized runtimes such as XLA, TensorRT, or fused-MLP kernels on modern GPUs/TPUs.

\section{Conclusion and future work}
\label{sec:conclusion}

In conclusion, this work introduced KD-PINN, a distillation-based framework that enhances the latency of PINNs by enabling fast-inference surrogates.
Besides acceleration, the study revealed the regularizing effect of knowledge distillation, that stabilizes the optimization landscape of PINNs, which is often ill-conditioned.

In terms of performance, KD-PINN transitions standard PINNs from near real-time inference ($\sim$20-50 ms for PINNs depending on the CPU) to the ultra-low-latency regime ($\sim$5 ms for KD-PINN distilled surrogates), defined by sub-10 ms responsiveness in context of control and simulation tasks.

The framework also allows the deployment on compute-limited embedded hardware where dedicated GPUs are impractical.
Potential applications include aerodynamics and CFD for rapid prediction of flow fields, real-time trajectory correction in robotics or embedded control, and instant pricing in computational finance.   Although distillation adds 20-25\% training cost due to teacher training, the resulting distilled models offer a substantial advantage in cases of repeated deployment or latency-critical scenarios.  

Future directions involve improving self-reliance in regard to the training parameters of the teacher and the self-tuning capability through self-distilling or meta-tuned architectures, and approaching theoretical acceleration limits via hardware-aware co-optimization. Finally, although the framework has been validated on a range of nonlinear PDEs with controlled dynamics, its behavior in strongly chaotic regimes remains to be investigated, for instance through the interaction between distillation, stability, and error propagation. In essence, KD-PINN aims to contribute to the development of scientific systems that learn and react as fast as the physics they emulate.

\bibliographystyle{unsrtnat}
\bibliography{KD-PINN4}

\end{document}